\documentclass[lettersize,journal]{IEEEtran}
\usepackage{amsmath,amsfonts}
\usepackage{algorithmic}
\usepackage{algorithm}
\usepackage{array}
\usepackage[caption=false,font=normalsize,labelfont=sf,textfont=sf]{subfig}
\usepackage{textcomp}
\usepackage{soul}
\usepackage{stfloats}
\usepackage{url}
\usepackage{verbatim}
\usepackage{graphicx}
\usepackage{cite}
\usepackage{xcolor}
\usepackage{amsmath}
\usepackage{amsthm}
\usepackage{amssymb}
\usepackage{lipsum}
\usepackage{multirow}
\usepackage{arydshln}
\usepackage{float}
\usepackage{etoolbox}
\newtheorem{prop}{Proposition}
\newcommand{\hquad}{\hspace{0.5em}} 
\usepackage{mathtools,amssymb,lipsum}

\usepackage{cuted}
\newtheorem{ass}{Assumption}
\newtheorem{Lemma}{\rm \textbf{Lemma}}

\newtheorem{innercustomthm}{Theorem}
\newenvironment{customthm}[1]
{\renewcommand\theinnercustomthm{#1}\innercustomthm}
  {\endinnercustomthm}

\begin{document}

\title{Straggler-resilient Federated Learning: Tackling Computation Heterogeneity with Layer-wise Partial Model Training in Mobile Edge Network}

\author{Hongda Wu,~\IEEEmembership{Student Member, IEEE},
        Ping Wang,~\IEEEmembership{Fellow,~IEEE}, C V Aswartha Narayana
\thanks{Hongda Wu, Ping Wang, and C V Aswartha Narayana are with the Department of Electrical Engineering and Computer Science, Lassonde School of Engineering, York University, Toronto, ON M3J 1P3, Canada (e-mail: hwu1226@yorku.ca; pingw@yorku.ca; cvan2410@yorku.ca, Work was done when Aswartha was a Mitacs intern at York University.)}

\thanks{A preliminary version of this work has been presented at the IEEE International Conference on Communications in China (ICCC), 2023 \cite{10233400}.}

}



\maketitle

\begin{abstract}
Federated Learning (FL) enables many resource-limited devices to train a model collaboratively without data sharing. However, many existing works focus on model-homogeneous FL, where the global and local models are the same size, ignoring the inherently heterogeneous computational capabilities of different devices and restricting resource-constrained devices from contributing to FL. In this paper, we consider model-heterogeneous FL and propose Federated Partial Model Training (FedPMT), where devices with smaller computational capabilities work on partial models (subsets of the global model) and contribute to the global model. Different from Dropout-based partial model generation, which removes neurons in hidden layers at random, model training in FedPMT is achieved from the back-propagation perspective. As such, all devices in FedPMT prioritize the most crucial parts of the global model. Theoretical analysis shows that the proposed partial model training design has a similar convergence rate to the widely adopted Federated Averaging (FedAvg) algorithm, $\mathcal{O}(1/T)$,  with the sub-optimality gap enlarged by a constant factor related to the model splitting design in FedPMT. Empirical results show that FedPMT significantly outperforms the existing benchmark FedDrop. Meanwhile, compared to the popular model-homogeneous benchmark, FedAvg, FedPMT reaches the learning target in a shorter completion time, thus achieving a better trade-off between learning accuracy and completion time.
\end{abstract}


\section{Introduction}
Federated Learning (FL) aims to learn a statistical model with good generalization capability at a large scale. Under the orchestration of an edge server, multiple edge devices participate in the FL process to achieve data augmentation while keeping training data locally \cite{FedAvg}. Without accessing user-sensitive data, user privacy is protected, and the communication cost is reduced, making FL one of the most promising technologies for future network intelligence \cite{park2019wireless}. FL process is divided into rounds of communication, each of which includes model distribution, local model training, and global model aggregation. Different from distributed machine learning, the learning process in FL poses a greater challenge due to the uncertainty of wireless networks and limited wireless resources, statistical heterogeneity, and system heterogeneity\footnote{FL is designed to be implemented within many of candidate devices typically equipped with different hardware capacities, e.g., CPU cycle, memory, and power, which is referred to as system heterogeneity. Further, training samples on each device are generated with user preference, which follows a heterogeneous distribution across different devices, termed statistical heterogeneity. }, making communication efficiency a critical barrier in FL implementation. 


To reduce the communication cost, methods including quantization \cite{quantizatino}, compression/sparsification \cite{compression, sparsification}, and over-the-air transmission \cite{overtheair} have been proposed. These techniques are dedicated to the wireless link where the transmission cost can be directly decreased by either sending a set of essential parameters or exploiting the superposition property of the wireless medium. Another perspective to alleviate the communication burden is improving the convergence speed (by reducing the communication rounds). To improve the slow convergence of FL with statistical heterogeneity, existing works proposed to use variance reduction \cite{scaffold}  participant selection \cite{BiasedSelection, PNS, 9796935}, important-based updating \cite{Adp, FirstorderP, SPcoding}, and clustering \cite{briggs2020federated, IFCA}. For example, studies \cite{BiasedSelection, PNS} choose devices according to their contribution to the global model. FedFOMO in \cite{FirstorderP} weighs model updates based on the loss values, which reflect the learning progress on local devices. IFCA \cite{IFCA} clusters participating devices into different groups by identifying device similarities, thus alleviating the impact of data heterogeneity. Meanwhile, deploying FL for real scenarios needs to consider the system heterogeneity, where device computation and/or communication capability, transmission uncertainty, and even the degree of participation could be different. For instance, authors in \cite{nishio2019client} choose participants based on the resource condition of local devices. The proposed method faces a bias and fair concern. Similarly, studies \cite{9488679} and \cite{9261995} focus on resource allocation problems and design FL systems considering energy consumption and learning time. Authors in \cite{amiri2021convergence} and \cite{chen2021communication} consider the impact of wireless channel on the convergence speed. 

It’s worth noting that an implicit assumption of a homogeneous model has been made in the above works, which encounters two fundamental challenges: 1) Device heterogeneity is a more realistic consideration. Since FL is designed to empower data information from massive devices, different types of devices are expected to be involved. Computation-heterogeneous devices make the model learning process problematic, especially in a federated network where the network identities (e.g., Internet of Things devices, PCs, and mobile devices) have different computational capabilities. Devices may only be capable of training models with capacities that match their on-device resources. 2) Machine learning has moved towards large models. Many machine/deep learning tasks rely on the use of large models\cite{bommasani2021opportunities}, for example, ResNet \cite{resnet}, Transformer\cite{transformer}. It is unrealistic to fit such large models to resource-constrained edge devices. 

To accommodate different types of devices with heterogeneous computational capabilities, \textit{model-heterogeneous} FL has been proposed in the research community. In this approach, participants are allowed to train models with different complexity (i.e., a subset of the whole model or even models with different architectures). To tackle the primary challenge of model aggregation, Knowledge Distillation \cite{hinton2015distilling} (KD)-based approaches have been proposed \cite{lin2020ensemble, cho2021personalized, he2020group, FedGen}. 
However, successful knowledge transfer and competitive results can only be achieved with desired public datasets, which are not always available in practice. 
Another series of works focuses on Partial Model Training (PMT)-based methods, which originated from concepts including model pruning \cite{FedMP, JiangIBM, 9835327} and dropout \cite{dropout, diao2021heterofl, fjord, FedRolex}. The main idea is that the server assigns sub-models (by random \cite{dropout, FedMP} or in a fixed way \cite{diao2021heterofl, JiangIBM}) to match the limited resource of local devices. The global/whole model is updated on the server side by aggregating sub-models from participants. Existing works \cite{FedMP, JiangIBM, dropout, diao2021heterofl, fjord, FedRolex, yu2018slimmable, yu2019universally, yuan2019distributed} generate sub-models by extracting the subset of each layer (e.g., in a neural network). From a statistical perspective, though each part of neural network layers has a chance to update, the global model is not evenly updated since FL training consists of multiple rounds of computation and aggregation. 
Additionally, when training a smaller model (for instance, a sub-model with only 20\% of the complexity of the full model) to accommodate participants with limited computational resources, it can introduce bias into the locally trained model, rendering the global model susceptible to the influence of biased sub-models.

In general discriminative model learning, the learning process can be viewed as feature extraction and classifier refinement, where shallow and deep features are learned. Authors in \cite{ClassifierCalibration} emphasize that accuracy is closely related to the classifier instead of shallow layers. A similar phenomenon is observed in the meta-learning context \cite{ANIL}. Empirically, a more significant bias in the classifier than in other layers is found in FL. Motivated by \cite{ClassifierCalibration}, in this paper, we try to answer the following questions: \textit{given limited computation on local devices, which part of the training model should be updated or protected in FL? Moreover, how does the server generate and assign sub-models to computation-heterogeneous devices?} As such, we propose a new method, \texttt{Fed}erated \texttt{P}artial \texttt{M}odel \texttt{T}raining (\texttt{FedPMT}), to generate sub-models in a layer-wise way for computationally heterogeneous devices to reduce the FL completion time. Different from existing PMT-based methods, which generate sub-models by preserving a subset of neurons in each layer  \cite{FedMP, JiangIBM, dropout, diao2021heterofl, fjord, FedRolex, yu2018slimmable, yu2019universally, yuan2019distributed}, FedPMT constructs sub-models from the back-propagation (BP) perspective. For resource-constrained devices, the computation burden is reduced by restricting gradient information from back-propagating to the shallow layers. Meanwhile, the most important layers (deep layers) are updated by back-propagation, and the local information (from each participant with unique data samples) is preserved in the partial model training process. 

To the best of our knowledge, this is the first work considering layer-wise model update to handle the system heterogeneity problem in FL\footnote{It is worth mentioning that a similar concept, sparsified BP \cite{sun2017meprop}, where only a small amount of parameters are updated in BP, is applied to reduce the over-fitting problem. Our paper is orthogonal to \cite{sun2017meprop} from both objective and implementation perspectives. }.
The primary advantage of our proposed FedPMT is that it enables all participants to prioritize the most crucial parts of the global model (i.e., deep layers) and ensure that local training achieves the purpose of data augmentation, as pursued in FL. Meanwhile, by avoiding the removal of neurons in deep layers, it guarantees a relatively large model capacity.

Our main contributions in this paper are as follows:
\begin{itemize}
    \item We identify the prospect of model-heterogeneous FL and propose a layer-wise partial model training strategy, FedPMT, for resource-constrained FL systems. In this paper, the proposed FedPMT accommodates heterogeneous computation over the FL system by counteracting back-propagation cost when updating the model, a.k.a. layer-wise. Without invoking further local computation overhead, FedPMT is an easy-to-implement framework, fully compatible with existing FL systems and secure aggregation protocols for privacy enhancement.
    
    \item We analyze the convergence property of the proposed design. FedPMT converges to the global optimum at a rate of $\mathcal{O}(1/T)$ for strongly convex and smooth function in data heterogeneous scenarios, which is similar to the FedAvg cases with no resource constraints. However, given heterogeneous computational capabilities on devices, FedPMT has a shorter task completion time.

    \item We empirically evaluate the performance of FedPMT via extensive experiments using the synthetic dataset and real datasets with different learning objectives. By analyzing the computation for various heterogeneous settings, our results demonstrate that the proposed design outperforms model-homogeneous (FedAvg) and model-heterogeneous (FedDrop) benchmarks regarding task completion time and training accuracy, respectively.
\end{itemize}

\section{Related Work}
In the realm of model training involving multiple parties, prior approaches have commonly adopted the concept of a slimmable neural network, where the target model is divided into different components, which are then trained in a distributed manner. In FL scenarios, the participating devices are naturally computation-heterogeneous; it is not trivial to consider model-heterogeneous FL design where the server assigns different models that match the devices' capabilities. To aggregate model information, existing literature can be categorized into two main streams: knowledge distillation and partial model training. Before diving into the FL context, we summarize a series of works focusing on subnet training to achieve a trade-off between accuracy and latency for model training\cite{yu2018slimmable, yu2019universally, yuan2019distributed}. 

\textbf{Slimmable Neural Network}: Yu \textit{et al.}  \cite{yu2018slimmable} propose to train a Slimmable Neural Network, i.e., several model variants (with a switch to control the model width) where the parameters on different variants are shared, and their individual information is kept by individual batch normalization layers. Further, authors in \cite{yu2019universally} propose Universal Slimmable Networks (US-Nets), which makes slimmable neural networks more generalized with any model width. Both Slimmable Neural Network and US-Nets aim to train several models simultaneously. In contrast, Yuan \textit{et al.}  \cite{yuan2019distributed} adopt the idea of individual subnet training called Independent Subnet Training (IST), where a large neural network is evenly divided into non-joint subnets, which are updated separately on different devices. IST focuses on cases where communication/memory is limited on a single device. Since no synchronization is required during local updates, per-step communication volume on multiple fronts can be reduced. Generally, the model in slimmable neural network \cite{yu2018slimmable, yu2019universally, yuan2019distributed} is evenly split into different devices. Though the flexibility to match devices' resources is achieved, the applicable scenario is limited. Authors in \cite{mask_convergent} analyze the convergence rate of model training with partial gradient by considering a more general setting where a partial of the neural network model is masked from updating. 

\textbf{Knowledge Distillation \cite{lin2020ensemble, cho2021personalized, he2020group, FedGen}}: One primary technique to exchange information between differently structured models is knowledge distillation (KD), where devices with less computational capability extract information from large models \cite{hinton2015distilling}. In FL, KD is used to transfer knowledge for both homogeneous models \cite{lin2020ensemble, cho2021personalized} and heterogeneous models \cite{he2020group}. Notably, FedDF proposed by Lin \textit{et al.} \cite{lin2020ensemble} 
initiates the training process by first training several classifiers on local private data. Subsequently, these local classifiers are employed to process public unlabeled data and generate logits. Each device's information is treated as a  ``teacher'' whose information is aggregated into the global model (``student'') to improve its generalization capability. Similarly, authors in  \cite{cho2021personalized} propose a cluster-based knowledge transfer where each device sends the logits (generated from local public data) back to the server, in which devices can be grouped, and local models are aggregated to different global models based on similarity. With multiple clusters, the proposed method is beneficial to alleviate the data heterogeneity problem. Note that local models in COMET \cite{cho2021personalized} are not necessarily homogeneous. FedGKT \cite{he2020group} extends KD to heterogeneous model scenarios. Local devices update the lightweight models in each global round. The information is then transmitted, aggregated, and incorporated into a large model on the server side via KD. The server's model could be larger than any local model. Meanwhile, with soft labels from server-side training, local models' performance is boosted by adopting the KD-based loss. To remove the dependence of public data in KD-based FL, authors in \cite{FedGen} propose to learn a generative model at the server side, which is solely derived from the output of local devices. Given target labels in the local side, the learned generator yields a feature representation consistent with the ensemble of each device's output. Though the generator provides information from other peer devices, transferring the generator is necessary in each round, incurring more communication costs.

\textbf{Partial Model Training in FL}: Authors in \cite{SPcoding} introduce the US-Nets to the FL context for the first time and propose superposition coding and successive decoding (for model transfer in uplink/downlink communication) to protect different parts of the learning model. In \cite{SPcoding}, only $0.5 \times$ model width (left/right model) is considered, and the learning performance is improved when including the partial models. Study \cite{SPcoding} focuses on the model transmission but requires local devices to train the whole model, which is impractical for devices with heterogeneous computational capabilities. Authors in \cite{diao2021heterofl} propose a computation and communication-efficient FL design for heterogeneous devices by allowing local models to have different architectures from the global model. Different from \cite{SPcoding, yu2018slimmable, yu2019universally}, the proposed \textit{HeteroFL} \cite{diao2021heterofl} grants local devices various model architectures (size) according to their computational capabilities and allows weak devices in terms of computation/ communication to contribute to the global model in FL. HeteroFL enables sub-model generation in a static way where sub-models are extracted from a designated part of the global model. Inspired by Dropout \cite{hinton2015distilling} in centralized machine learning, it is straightforward to adopt Dropout to FL to alleviate the resource-constrained local computing. As illustrated in \cite{dropout}, the FL server randomly removes a subset of neurons and generates sub-models for the participating device to meet its computation level. Similar to the static sub-model generation as in \cite{diao2021heterofl}, authors in \cite{fjord} propose FjORD, which combines the sub-model training and knowledge distillation to improve the sub-model performance. As stated in \cite{FedRolex}, both \cite{diao2021heterofl, dropout, fjord} suffer from performance degradation on high data heterogeneity. This is primarily due to the limitation that different sub-models can only be updated on specific devices that match their computation level, forcing different parts of the global model to be updated on samples with different distributions. In addition, performance degradation in Federated Dropout is related to the randomness in the device cohort when generating a partial model. To overcome the drawback of random \cite{dropout} and static \cite{fjord, diao2021heterofl} partial-model generation, authors in \cite{FedRolex} propose using a rolling window to counteract uneven model updates by which all parts of the global model are looped in sequence. This rolling process iterates each round until the global model is evenly trained to converge. 

Overall, works in \cite{yu2018slimmable,yu2019universally, yuan2019distributed, SPcoding,diao2021heterofl} face covariant shift problems because the partial model is constructed by statically downsampling in each layer. Therefore, different sub-models can only be trained on specific devices that match the resource constraint, updating different parts of the global model with different data distributions. This drawback would degrade the training performance, especially in data heterogeneous FL scenarios. Since the expectation of the output feature in the partial model differs from that in the full model, one must add batch normalization layers or manually scale the output feature. Authors in \cite{FedRolex} propose to handle the problem with a rolling window where different sub-models (in each layer) can be updated more evenly. Even though sub-model generation is updated in each global round, multiple local updates during consecutive rounds might cause skewness in model learning. More importantly, all the works mentioned above discard a specific ratio of \textit{weight} (the connection between neurons in layers) in every layer to generate a sub-model, which might not be necessary for model training in FL since features in shallow layers are less important and can be shared among devices while unique features of devices are revealed by keeping (at least) the classifier updated. Our proposed layer-wise partial model training strategy inaugurates a new direction of handling computation heterogeneity in FL due to its effectiveness, simplicity, and scalability.

\begin{table*}[t]
\centering
\caption{Comparisons of model-homogeneous and model-heterogeneous FL design in the existing literature and the proposed FedPMT}
\label{tab:requirement_comparison_methods}
\resizebox{1.0\linewidth}{!}{%
\begin{tabular}{lccccccc} 
\multicolumn{2}{c}{ FL Methods}  & \begin{tabular}[c]{@{}c@{}} {Model}\\ { Heterogeneity}\end{tabular} & \begin{tabular}[c]{@{}c@{}} {Aggregation}\\ {Scheme}\end{tabular} & \multicolumn{1}{l}{\begin{tabular}[c]{@{}l@{}} {Comp. / Comm. }\\ { Heterogeneity}\end{tabular}} &  \begin{tabular}[c]{@{}c@{}} {Sub-model Generation /}\\ {Auxiliary Data}\end{tabular}  \\ 
\hline
\multirow{2}{*}{\begin{tabular}[c]{@{}c@{}} Convergence \\ Optimization \end{tabular} }  & FedAvg \cite{FedAvg} & \multirow{2}{*}{No} &   \multirow{2}{*}{-}
&  \multirow{2}{*}{- / No} & \multirow{2}{*}{- / No}\\
& SCAFFOLD \cite{scaffold} &  &   & &  \\
\hline
\multirow{4}{*}{\begin{tabular}[c]{@{}c@{}} Knowledge \\ Transfer \end{tabular} } & FedDF  \cite{lin2020ensemble } & \multirow{4}{*}{Yes} & \multirow{4}{*}{\begin{tabular}[c]{@{}c@{}}Knowledge \\Distillation \end{tabular}} & \multirow{4}{*}{-} & - / Unlabeled \\
& COMET \cite{cho2021personalized} &  &  &  & - /  Unlabeled  \\
& FedGKT \cite{he2020group} &  &  &  & - / No  \\
& FedGen \cite{FedGen} &  &  &  & - / No (Generator)   \\
\hline
\multirow{2}{*}{\begin{tabular}[c]{@{}c@{}} Model \\ Prunning \end{tabular} }  & FedMP  \cite{FedMP } & \multirow{2}{*}{Yes} & \multirow{2}{*}{\begin{tabular}[c]{@{}c@{}} - \end{tabular}} & $\checkmark$ / $\checkmark$  & Random / No  \\
& PruneFL \cite{JiangIBM} &  &  & $\checkmark$ / $\checkmark$ & Static / No  \\
\hline
\multirow{6}{*}{\begin{tabular}[c]{@{}c@{}} Partial Model \\ Training \end{tabular} } & Federated Dropout \cite{dropout} & \multirow{6}{*}{Yes} & \multirow{6}{*}{\begin{tabular}[c]{@{}c@{}}Sub-model \\Training\end{tabular}} & $\checkmark$ / $\checkmark$ & Random / No \\
& HeteroFL \cite{diao2021heterofl} &  & &  $\checkmark$ / - & Fixed / No  \\
& SlimFL \cite{SPcoding}  &  & &  \quad - / $\checkmark$  & Fixed / No  \\
& FjORD \cite{fjord} &  &  &  $\checkmark$ / $\checkmark$ & Ordered /  No \\
& FedRolex \cite{FedRolex} &  &  & $\checkmark$ / $\checkmark$ & Rotated /  No \\ \cdashline{2-6} \\
& \textbf{FedPMT (Ours Approach)} &  &  & $\checkmark$ / $\checkmark$ & \textbf{layer-wise} /  No \\
\hline
\end{tabular}
}
\vspace{-4mm}
\end{table*}


\section{Preliminary}
\subsection{Federated Learning}

We consider a federated network that includes one central server and a set of local devices $\mathcal{K}$ with size denoted by $|\mathcal{K}|$ (we use the Cardinality of a set to represent its size hereinafter). The goal in FL is to learn a parametric model $\mathbf{w}$ that fits data samples in a distributed setting by minimizing loss function $F(\mathbf{w})$. In particular, we assume each local device $k \in \mathcal{K}$ has a training set $\mathcal{D}_k$ that follows a data distribution $q_k$, i.e., each sample $z_{k,1}, z_{k,2} \cdots z_{k,|\mathcal{D}_k|}$ is drawn from $q_k$ distribution randomly, where each sample consists of a pair of feature and response denoted by $z_{k,s} = \{x_{k,s}, y_{k,s}\}$. Let $\ell (\mathbf{w}; z_{k,s}): \Theta \rightarrow \mathbb{R}$ be the loss function associated with data sample $z_{k,s}$, where $\Theta = \mathbb{R}^d $ is the parameter space. The population loss function for each device $k$ is defined as
$F_k( \mathbf{w} ):= \mathbb{E}_{z_{k,s} \backsim q_k} \left[ \ell (\mathbf{w}; z_{k,s}) \right] $. Because each device has a small number of data samples, population distribution on the device is not fully observed. Instead of minimizing the population loss function, each device targets the Empirical Risk Minimization (ERM) problem defined as
\begin{align*}
\label{eq1}
F_k( \mathbf{w} )= \frac{1}{|\mathcal{D}_k|} \sum_{z_{k,s}\in \mathcal{D}_k} \ell (\mathbf{w}; z_{k,s}). \tag{1}
\end{align*}

The FL objective is to minimize a surrogated function
\begin{align*}
\label{eq2}
\underset{\mathbf{w}}{\min} \ F( \mathbf{w} ) :=  
 \sum_{k =1}^{|\mathcal{K}|}  \frac{|\mathcal{D}_k|}{\sum_{k =1}^{|\mathcal{K}|} |\mathcal{D}_k| }  F_k( \mathbf{w} ). \tag{2}
\end{align*}

A canonical way of solving the above objective is Federated Averaging (FedAvg) \cite{FedAvg}, which is a variant of Stochastic Gradient Descent (SGD) with multiple (global) rounds, where each round consists of multiple steps of local update (e.g., $\tau$ steps) followed by model synchronization process between participating devices and the server. Denoting $t = \{1,2, \cdots, T\}$ as the index of FL global rounds, one round of FedAvg is described as
\begin{enumerate}
\item The server selects a subset of devices $\mathcal{S} \subseteq \mathcal{K}$ uniformly at random and broadcasts the latest model $\mathbf{w}^{t}$ to the chosen devices $k \in\mathcal{S} $.
\item Each selected device views $\mathbf{w}^{t}$ as an initial and updates it by $\tau$ steps of SGD over its empirical risk objective defined in (1), and sends $\mathbf{w}_k^{t+1} $ back to the server.
\item The server aggregates received local models $\mathbf{w}_k^{t+1}, k\in \mathcal{S}$ with weight  $ |\mathcal{D}_k| / \sum_{k =1}^{|\mathcal{S}|} |\mathcal{D}_k|  $ and gets model $\mathbf{w}^{t+1}$.
\end{enumerate}

The above steps repeat until a satisfying learning result, e.g., the learning accuracy in classification tasks, is met.

\subsection{System Heterogeneity}
Several works have demonstrated the effectiveness of FedAvg from both empirical and theoretical perspectives in various settings \cite{Li2020On, haddadpour2019convergence, scaffold}. One needs to notice that in the system heterogeneous FL, the assumption that every participating device can timely train the designated model and/or transmit the updated model back to the server may not always hold true. For example, the network identities in heterogeneous networks can be Internet of Thing (IoT) devices, PCs, and mobile devices, which have different computational and/or communication capabilities. Devices may not be able to train a large model due to their energy consumption on this task, or their CPU cycles are too small to finish the task on time, causing a long delay or straggler effect \cite{fjord, 9139873}. Therefore, it is not trivial to design an FL system from the time consumption perspective and consider the system heterogeneity. In what follows, we introduce a computation model in general FL.

We denote the number of CPU cycles for device $k$ to execute one sample of data by $c_k$, which is considered a priori information and can be measured offline. Suppose that all samples $z_{k,s} \in \mathcal{D}_k$ have the same size (e.g., the number of pixels in images), the number of CPU cycles required by device $k$ for each time of local training (i.e., one global round) is $ c_k \cdot |\mathcal{D}_k| \cdot E$, where $E$ is the number of local training epoch. Furthermore, the computation time for each global round is derived as
$ T_{cmp}^t = \frac{c_k \cdot |\mathcal{D}_k| \cdot E}{\kappa_k}  $, where $\kappa_k$ is the CPU cycle frequency of device $k$, which is fixed for one device and varies for different devices. In this paper, the system heterogeneity is reflected by $\kappa_k$. This is because $c_k$ is a constant given a training model $\mathbf{w}$, so 
devices with a higher value of $\kappa_k$ signify a larger computational capacity, enabling them to complete the local training process faster. In a typical FL design, all devices are rehearsed with the same number of SGD steps (i.e., $\tau$ ). Therefore, devices with small computational capability would spend a long time to finish the local training, resulting in the straggler effect \cite{fjord, 9139873}. In this paper, we do not consider the convergence improvement by assigning adaptive $\tau$, which is determined by $|\mathcal{D}_k|$, $E,$ or SGD batch-size as in \cite{8664630, 9488679}, but focus on delivering different partial models to different participants to mitigate the impact of system heterogeneity. 
The time consumption for participating devices in each global round $t$ is bounded as $\max \{ \frac{c_k \cdot |\mathcal{D}_k| \cdot E}{\kappa_k} \}, k \in \mathcal{S}$.

\section{Partial Model Training}

\begin{figure*}[t!]
\centerline{\includegraphics[scale=0.4]{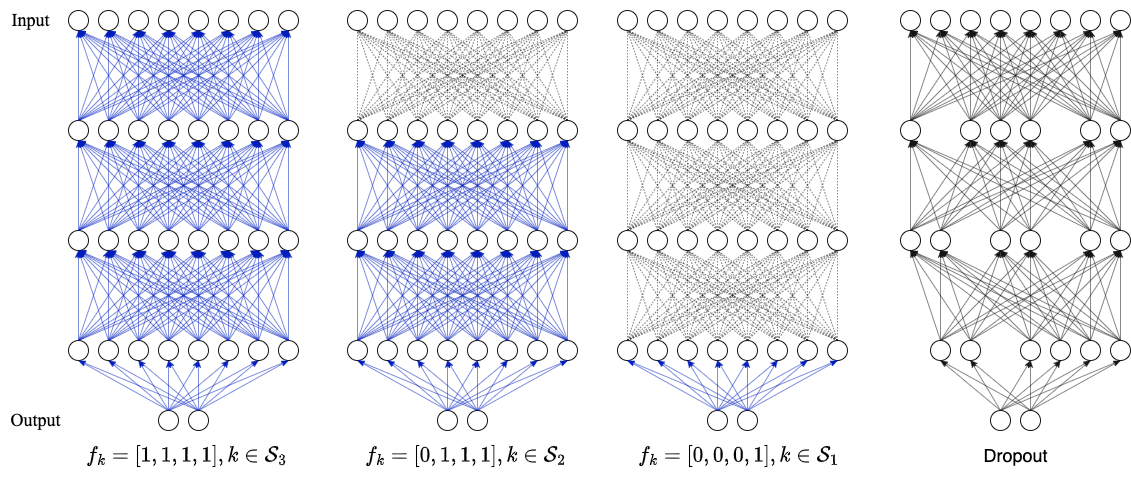}}
\caption{Illustration of different local training models of the 
proposed FedPMT for four layers of fully connected neural networks (i.e., $|\mathcal{L}| = 4$) with model width $\mathcal{I} = \{1,2,3\}$, where $|\mathcal{I}| = 3 < |\mathcal{L}| = 4$. The leftmost plot represents the model with full model width $|\mathcal{I}|=3$. The partial model training process with mask $\Xi_i, i \in \mathcal{I}$ is shown by dotted lines and arrow lines in blue, i.e., weights with arrow lines in blue are updated using BP. The weights with dotted lines are not updated by BP, where only the forward process is involved.
Function $f_k, k \in \mathcal{S}_i, i\in\mathcal{I}$ is given to represent $\Xi_i$ using $\Upsilon_k^l, l \in \mathcal{L}$. In comparison, FedDrop \cite{dropout} removes neurons in hidden layers with probabilities (to accommodate different computational capabilities on devices) at random. For example,  the model after dropout (with a dropout rate of 0.25) is shown in the rightmost plot.}
\label{system}
\end{figure*}

\subsection{System Model}
We consider an FL scenario where participating devices have heterogeneous computing capabilities. We adopt the concept of partial model training to accommodate the contribution of devices with heterogeneous computing capabilities. To better compromise the computation heterogeneity, the server provides a variety of computing options reflected by model width $\mathcal{I} = \{1,2, \cdots, |\mathcal{I}|\}$ to those devices for motivating participation and improving the global model convergence. Specifically, at each global round $t = 1,2, \cdots, T$, each participating device can choose one of the model widths provided in $\mathcal{I}$ for its local model training process according to its computing capability.

Without loss of generality, in the $t$-th global round, we use $|\mathcal{I}|$ to indicate the full model width, use $\Xi_i, i \in \{1,2, \cdots, |\mathcal{I}|\}$ as the \textit{mask} to generate the local model in order to do partial training and use non-joint sets $\mathcal{S}_i, i \in \{1,2, \cdots, |\mathcal{I}|\}$ to represent the corresponding sets (with the same model width) that devices belong to. As such, device $k\in \cup \mathcal{S}_i, i \in \mathcal{I}\setminus  \{|\mathcal{I}|\}$ can generate a partial model (based on its computing capability) as $\mathbf{w}_{k}^t = \mathbf{w}^t \odot \Xi_i$  for further processing, and it is clear that $\mathbf{w}_{k}^t = \mathbf{w}^t \odot \Xi_{|\mathcal{I}|} = \mathbf{w}^t$ holds for devices with full model width, i.e., $k\in \mathcal{S}_{|\mathcal{I}|}$. $\odot$ operated on model $\mathbf{w}^{t}$ is defined to represent the partial model generation process, which is illustrated as 
\begin{align*}
\label{eq4}
\mathbf{w}_{k}^t = 
    \begin{cases}
       \mathbf{w}^t \odot \Xi_i & k\in \mathcal{S}_i, i \in \mathcal{I}\setminus  \{|\mathcal{I}|\} \\          
 \mathbf{w}^t  \odot \Xi_i =  \mathbf{w}^t & k\in \mathcal{S}_{i}, i = |\mathcal{I}| 
    \end{cases} . \tag{4}
\end{align*}

In the proposed FedPMT, we achieve partial model training from the perspective of Back-Propagation (BP)\footnote{Our proposed scheme is different from existing works\cite{ yu2018slimmable,yu2019universally, yuan2019distributed}, which split the training model into different sub-models with overlap \cite{yu2018slimmable,yu2019universally} or without overlap \cite{yuan2019distributed}. However, these works mentioned above split the network from the neurons' perspective (by only including partial parameters of each layer of the training model, as seen in Fig. \ref{system}.). Our work splits the training model from the layers' perspective. Partial model training means that devices with model width $i \in \mathcal{I}\setminus  \{|\mathcal{I}|\}$ exclusively update part of layers of this model, from the back to the front. As with the traditional FL design, all the participating devices update the classifier, i.e., the last layer of the model, which is helpful to alleviate the classification bias that is identified as the culprit of FL with heterogeneous data \cite{ClassifierCalibration}.}. Particularly, all participating devices $k\in \cup \mathcal{S}_i, i\in \mathcal{I}$ share the same forward process, i.e., calculating the loss function given current model $\mathbf{w}^{t}$ and its data samples $z_{k,s} \in \mathcal{D}_k$. Differently, devices without full model width, i.e., $k\in \cup \mathcal{S}_i, i\in \mathcal{I} \setminus |\mathcal{I}|$ \textit{will not} update all the parameters in BP process, and only update the parts where BP is involved instead. This is achieved by restricting gradient information from back-propagating to the shallow layers. We introduce $\Upsilon_k^l, l \in \mathcal{L} = \{1,2, \cdots, |\mathcal{L}| \}$ to indicate whether the $l$-th layer of the learning model on device $k$ is involved in the BP process, where $|\mathcal{L}|$ is the total number of layers in the model. Therefore, for each device $k\in \mathcal{S}_i$ with mask $\Xi_i$, a relationship between $\Xi_i$ and $\Upsilon_k^{l}$ is generated $\Xi_i = f_k ( \sum_{l\in \mathcal{L}} \Upsilon_k^l), i \in \mathcal{I}$ to represent the involved layers in BP, where $\Upsilon_k^l$ is a vector with binary values. The $l$-th element of $\Upsilon_k^l$ is 1, indicating that the $l$-th layer of device $k$ is involved in BP; otherwise, all elements in $\Upsilon_k^l$ are 0. $f_k ( \sum_{l\in \mathcal{L}} \Upsilon_k^l) $ is regarded as a mapping function with binary coefficients that shows which layer's gradient is update \footnote{Suppose that a three-layer model $\mathbf{w}$ is divided into three different widths, i.e., $\mathcal{I} = \{1,2, 3\}$. With $i=3$ being the full model, i.e., $\mathbf{w} \odot \Xi_3 = \mathbf{w}$, devices within set $\mathcal{S}_3$ will update all layers using the BP process. In this case, $f_k$ is written as $[1,1,1]$ and $\Upsilon_k^1=[1,0,0] , \Upsilon_k^2=[0,1,0], \Upsilon_k^3=[0,0,1]$ (we remove the subscription $k$ for generalization). Similarly, for those devices $k \in \mathcal{S}_1$ with model width $\Xi_1$, $f_k$ is viewed as $[0,0,1]$ ($\Upsilon_k^1=[0,0,0] , \Upsilon_k^2=[0,0,0], \Upsilon_k^3=[0,0,1]$), which means that only the last layer is involved in the BP process, and the first two layers will not be updated by BP. In more general cases where $|\mathcal{I}|$ is less than the total number of model layers (e.g., $|\mathcal{I}| = 3 \leq |\mathcal{L}| = 4$), $f_k$ can be generated similarly, e.g., $f_k = [0,1,1,1]$ and $f_k = [1,1,1,1]$. Refer to Fig. \ref{system} for a detailed illustration. }. 


In what follows, notation $f_k ( \sum_{l\in \mathcal{L}} \Upsilon_k^l) $ is simplified as $f_{k} $. Similar to the vanilla federated optimization \cite{FedAvg}, each device minimizes its empirical risk as shown in (\ref{eq1}) by running $\tau$ steps of (mini-batch) SGD to update local parameters initialized as $\mathbf{w}^{t}$. For device $k$, the local model training is formally expressed as
\begin{align*}
\label{eq6}
\mathbf{w}_{k}^{t+1} = \mathbf{w}^{t} - \eta_t \underbrace{\nabla F_k(\mathbf{w}^{t}, \xi_k) \circ f_{k} }_{\tilde{\nabla F_k}(\mathbf{w}_k^{t}, \xi_k) },
    \tag{6}
\end{align*}
where $\eta_k$ is the learning rate, $\xi_k$ is the mini-batch samples, and $\tilde{\nabla F_k}(\mathbf{w}_k^{t}, \xi_k)$ is the actual gradient for model update in device $k$, which might be the partial or full gradient depending on the binary values of $f_{k}$.  We use $\circ$ to represent the \textit{layer-wise multiplication} between a vector $\mathrm{a}$ of length $|\mathcal{L}|$ and gradient vector $\mathrm{b}$ with $|\mathcal{L}|$ blocks/layers, Below shows a general example for layer-wise multiplication: $\mathrm{a} = [0, 1, 2], \mathrm{b} = [2, 2, 2; 1, 1; 3, 3, 3, 3]$. In the vector $\mathrm{b}$, the semicolon ‘;’ serves as a delimiter to distinguish between model parameters across different layers. We have $\mathrm{a} \circ \mathrm{b}$ = [0, 0, 0; 1, 1; 6, 6, 6, 6]. Note that the length of vector $\mathrm{a}$ equals the number of layers in gradient vector $\mathrm{b}$. As such, by introducing the mask, devices with model width $i \in \mathcal{I}\setminus  \{|\mathcal{I}|\}$ will not update the model weight of front layers, thus alleviating the computational burden (e.g., for partial derivative and matrix multiplication). 

\begin{figure*}
\begin{align*}
\label{eq7}
\mathbf{w}^{t+1} & 
\stackrel{(a)}= \mathbf{w}^{t} - \eta_t \sum_{k \in \mathcal{S}} A_k \circ \tilde{\nabla F_k}(\mathbf{w}_k^{t}, \xi_k) \\
& \stackrel{(b)} = \mathbf{w}^{t} - \eta_t \left( \frac{1}{|\mathcal{S}_{ |\mathcal{I}|}| } \sum_{k \in \mathcal{S}_{|\mathcal{I}|}} \nabla F_k(\mathbf{w}^{t}, \xi_k) \circ \Upsilon_k^1  + \frac{1}{|\mathcal{S}_{|\mathcal{I}|} \cup \mathcal{S}_{|\mathcal{I}|-1} |} \sum_{k\in \mathcal{S}_{i|\mathcal{I}|} \cup \mathcal{S}_{|\mathcal{I}|-1}} \nabla F_k(\mathbf{w}^{t},\xi_k) \circ \Upsilon_k^2 \right. \\ 
& \left. \quad + \cdots + \frac{1}{| \cup \mathcal{S}_{i,i \in \mathcal{I}} |} \sum_{k \in \cup \mathcal{S}_i, i\in \mathcal{I}} \nabla F_k(\mathbf{w}^{t} ,\xi_k) \circ\Upsilon_k^{|\mathcal{I}|} \vphantom{\frac{1}{|\mathcal{S}_{ |\mathcal{I}|}| }} \right) \\ 
& = 
\mathbf{w}^{t} - \eta_t \underbrace{\sum_{i \in \mathcal{I}}  \frac{1}{|\cup \mathcal{S}_{j, j=i,\cdots, |\mathcal{I}|} |} 
 \sum_{k \in \cup \mathcal{S}_j} \nabla F_k(\mathbf{w}^{t}, \xi_k) \circ  \Upsilon_k^{|\mathcal{I}|-i+1}}_{\nabla F(\mathbf{w}^{t})}, \tag{7} 
\end{align*}
\end{figure*}

After the local training, the server collects the model updates $\tilde{\nabla F_k}(\mathbf{w}_k^{t}, \xi_k), k\in \mathcal{S}$ to cast the global model, as shown in (\ref{eq7}), where $A_{k} = [a_1, a_2, \cdots, a_{|\mathcal{L}|}] $and $a_l = \frac{f_k [l]}{\sum_{k \in \mathcal{S}} f_{k} [l]}$ for $l=1,2,\cdots, |\mathcal{L}|$ and $k\in \mathcal{S}$, where $f_k [l]$ is the $l$-th element of $f_k$. For simplicity of representation, we consider the size of local datasets on local devices to be the same. Equation (\ref{eq7}) gives two different ways to represent global model aggregation, i.e., from the device's perspective (equation ($a$)) or the layer-wise perspective (equation ($b$)). As shown in ($a$), different from weighing local models with a single scalar \cite{FedAvg, Adp}, a weighting vector $A_k$ whose values correspond to the specific layer-wise weight in aggregation, is allocated to local models since devices may provide a partially updated model\footnote{The values in $A_k$ indicate that devices can provide partial gradient/model for aggregation, and parameters of the rest parts of the partial model are not counted in aggregation because no gradient update is done for those parameters.}. For example, $A_{k}[2]$ indicates the weight for aggregating the model parameters of the 2nd layer of device $k$. $\circ$ represents the layer-wise multiplication calculation between weighting vector $A_k$ and gradient $\tilde{\nabla F_k}(\mathbf{w}_k^{t}, \xi_k)$.

\begin{algorithm}[t]
  \caption{Federated Learning with Partial Model Training}
  \begin{algorithmic}[1]
 \STATE  \textbf{Input:} device set $\mathcal{K}$, step size $\eta_t$, model initialization $\mathbf{w}^1$ \\ number of global round $T$, number of local steps $\tau$, \\
$\kappa$ if Option I is chosen
  \FOR{$t= 1, \ldots, T $}
  \STATE \underline{Server:} $\mathcal{S}_t \leftarrow$ random subset of $\mathcal{K}$
  
  \STATE \textbf{Option I} (server initiates model splitting):
  \STATE \hspace{2mm}  send $\mathbf{w}^t$ and $f_{k}$ to device $k \in \mathcal{S}_i$ 
  \STATE \textbf{Option II} (device initiates model splitting):
  \STATE \hspace{2mm} send $\mathbf{w}^t$ and different masks $\Xi_i, i \in \mathcal{I}$ to device $k \in \mathcal{S}$

  \FOR{\underline{local device $k\in \mathcal{S}_t$} in parallel}
  
  \STATE \textbf{Option I} (server initiates model splitting):
  \STATE \hspace{2mm}   $\mathsf{LocalUpdate} ( \mathbf{w}^t, f_k, \eta_t, \tau)$
  \STATE \textbf{Option II} (device initiates model splitting): 
  \STATE \hspace{2mm} choosing appropriate $f_k = \Xi_i$ based on computational capability
  \STATE \hspace{2mm} $\mathsf{LocalUpdate} (\mathbf{w}^t, f_k, \eta_t, \tau)$
  \ENDFOR
  \STATE \underline{Server:} model aggregation by (\ref{eq7})
    \ENDFOR
\STATE \textbf{return} 

\underline{ $\mathsf{LocalUpdate} ( \mathbf{w}^t, f_k, \eta_t, \tau)$ } at the $k$-th device
  \FOR {$step = 1,\ldots,\tau $} 
  \STATE (mini-batch) stochastic gradient descent by (\ref{eq6})
  \ENDFOR
  \STATE \textbf{return} $\tilde{\nabla F_k}(\mathbf{w}_k^{t}, \xi_k)$
  \end{algorithmic}
  \label{alg1}
\end{algorithm}
Procedures of the proposed FedPMT algorithm are summarized in Algorithm \ref{alg1}. Two options are provided, as seen in Algorithm \ref{alg1}, where the mask for partial model training is generated either by the server or by local devices. If we choose option I, where the server
generates the mask, then each device should report its computation level to the server, similar to \cite{nishio2019client}. Otherwise, option II can be adopted, in which each device determines mask $f$ that matches its computation level. In this case, only marginal extra information $f$ is added to the uplink model transmission.

\subsection{Convergence Analysis}
In this section, we analyze the convergence property of the proposed partial model training with the local objective satisfying the strongly convex and smooth assumptions and compare it to the convergence rate of FedAvg. For the ease of theoretical analysis, we consider $|\mathcal{I}| = |\mathcal{L}|$ and scenarios with $|\mathcal{I}| \leq |\mathcal{L}|$ are verified in Section \ref{resultC}. To facilitate the convergence analysis, we also introduce assumptions \ref{ass3} and \ref{ass4}, which are commonly adopted in the literature \cite{Li2020On, SPcoding, stich2018local}.
\begin{ass} 
\label{ass1}
\textbf{$\mu$-strong convexity}. $F_k( \mathbf{w}), k\in \mathcal{S}$ is $\mu$-strong convex, i.e. $F_k( \mathbf{w}) \geq F_k( \mathbf{w'}) + (\mathbf{w} - \mathbf{w'})^\top \nabla F_k(\mathbf{w'}) + \frac{\mu}{2} \Vert \mathbf{w} - \mathbf{w'} \Vert^2 $, for all $\mathbf{w}, \mathbf{w'}$, where $(\cdot)^\top$ denotes the transpose operation of vector.
\end{ass} 

\begin{ass} 
\label{ass2}
\textbf{$L$-smoothness}. $F_k( \mathbf{w}), k\in \mathcal{S}$ is $L$-smooth, i.e. $F_k( \mathbf{w}) \leq F_k( \mathbf{w'}) + (\mathbf{w} - \mathbf{w'})^\top \nabla F_k(\mathbf{w'}) + \frac{L}{2} \Vert \mathbf{w} - \mathbf{w'} \Vert^2$, for all $\mathbf{w}, \mathbf{w'}$.
\end{ass} 

\begin{ass} 
\label{ass3}
\textbf{Bounded local gradient}. The expected squared norm of the local stochastic gradient is bounded,\\
i.e., $\mathbb{E}  \Vert \nabla F_k(\mathbf{w}_{k}^{t}, \xi_k) \Vert^2 \leq G^2$ for all device $k \in \mathcal{S}$ and $t = 1,2,\cdots, T$.
\end{ass}

\begin{ass} 
\label{ass4}
\textbf{Bounded local gradient variance}. The variance of local gradient $\nabla F_k(\mathbf{w}_{k}^{t}, \xi_k)$ is bounded,\\
i.e., $\mathbb{E} \left[ \Vert \nabla F_k(\mathbf{w}_{k}^{t}, \xi_k) - \nabla F_k(\mathbf{w}_{k}^{t}) \Vert^2
\right] \leq \delta_k^2$, with $\nabla F_k(\mathbf{w}_{k}^{t}) $ denoting the ground-truth gradient over device $k$ given $\mathcal{D}_k$. 
\end{ass}

Similar to \cite{8664630}, we define $\delta^2 = \frac{1}{|\mathcal{S}|} \sum_{k\in \mathcal{S}} \delta_k^2$ to measure the overall data heterogeneity of all devices in federated optimization. Please note that Assumptions \ref{ass1} and \ref{ass2} are commonly considered in FL analysis \cite{Li2020On, fedprox, stich2018local}, e.g., the learning objective can be logistic regression and softmax classifier with $\ell_2$ norm. Assumptions \ref{ass3} and \ref{ass4} have been made in previous works \cite{Li2020On, SPcoding, stich2018local}. 


\begin{prop}
Given local loss satisfying $\mu$-strong convexity, the following inequality can be derived, i.e., $\langle \mathbf{w}_k^{t} - \mathbf{w}^{*}, A_k \circ F_k(\mathbf{w}_k^{t}) \rangle \geq \frac{\varepsilon}{|\mathcal{S}|}  \cdot \left( F_k(\mathbf{w}_k^{t}) - F_k(\mathbf{w}^*) + \frac{\mu}{2} \Vert \mathbf{w}_k^{t} - \mathbf{w}^* \Vert^2  \right) $ for devices $k\in \mathcal{S}$ in FedPMT, where $\varepsilon \in [0,1]$ indicates the information loss due to the partial model update.
\end{prop}

\begin{proof}
Given the loss function satisfying Assumption \ref{ass1}, $\langle \mathbf{w}_k^{t} -  \mathbf{w}^{*}, \nabla  F_k(\mathbf{w}_k^{t}) \rangle \geq F_k(\mathbf{w}_k^{t}) - F_k(\mathbf{w}^*) + \frac{\mu}{2} \Vert \mathbf{w}_k^{t} - \mathbf{w}^* \Vert^2 $, Proposition 1 is derived based on the fact that all parts (reflected by each layer of gradient $\nabla F_k(\mathbf{w}_k^{t})$) of the model contribute to the local objective minimization (i.e., the right-hand side of above inequality), and removing part of the model information results in the slowness of the minimization process.

We assume the \textit{most} information decrement on devices with partial model update is measured by $1-\varepsilon$, and whose bound, i.e., $\left( F_k(\mathbf{w}_k^{t}) - F_k(\mathbf{w}^*) + \frac{\mu}{2} \Vert \mathbf{w}_k^{t} - \mathbf{w}^* \Vert^2  \right)$ is lowered by a constant factor $\varepsilon$ in such cases. 
The rationality behind the reduced bound related to the constant factor $\varepsilon$ reveals that these devices can retain at least the following amount of information $ \varepsilon \cdot   \left( F_k(\mathbf{w}_k^{t}) - F_k(\mathbf{w}^*) + \frac{\mu}{2} \Vert \mathbf{w}_k^{t} - \mathbf{w}^* \Vert^2  \right)$, though they update the model with the least effort due to the computation constraint. Since devices $k\in \mathcal{S}_{1}$ only update the last layer of the model, thus lose the most information regarding its local objective minimization process. Consequently, with weighting vector $A_k, k\in \mathcal{S}_{1}$ for aggregation being $[0,\cdots, \frac{1}{|\mathcal{S}|}]$, the following inequality is achieved, i.e., $\langle \mathbf{w}_k^{t} -  \mathbf{w}^{*}, A_k \circ \nabla F_k(\mathbf{w}_k^{t}) \rangle \geq  \frac{\varepsilon}{|\mathcal{S}|}\left( F_k(\mathbf{w}_k^{t}) - F_k(\mathbf{w}^*) + \frac{\mu}{2} \Vert \mathbf{w}_k^{t} - \mathbf{w}^* \Vert^2  \right), \forall k \in \mathcal{S}_1$. For all the other devices $k\in {\cup \mathcal{S}_{i, i=2,\cdots, |\mathcal{I}|}}$ that update more layers in partial model training and thus can retain more information, the above inequality is fulfilled. Therefore, the inequality is achieved for all devices in partial model training scheme, i.e., $\langle \mathbf{w}_k^{t} -  \mathbf{w}^{*}, A_k \circ \nabla F_k(\mathbf{w}_k^{t}) \rangle \geq  \frac{\varepsilon}{|\mathcal{S}|} \left( F_k(\mathbf{w}_k^{t}) - F_k(\mathbf{w}^*) + \frac{\mu}{2} \Vert \mathbf{w}_k^{t} - \mathbf{w}^* \Vert^2  \right), \forall k \in \mathcal{S}$.
\end{proof}

\begin{Lemma}
\label{lem1}
\textbf{(Bounded variance for global gradient)}. From Assumption \ref{ass4}, the variance of global gradient is bounded as $\mathbb{E} \left[ \Vert \nabla F(\mathbf{w}^{t}) - \nabla \bar F(\mathbf{w}^{t}) \Vert^2 \right] \leq 2 |\mathcal{I}|\delta^2 \psi$, where $\nabla F(\mathbf{w}^{t})$ and $\nabla \bar F(\mathbf{w}^{t})$ represent the global gradient surrogated by $\nabla F_k(\mathbf{w}_{k}^{t}, \xi_k)$ and $\nabla F_k(\mathbf{w}_{k}^{t})$, respectively, according to the aggregation method in (\ref{eq7}). $\psi =   \sum_{i \in \mathcal{I}}  \frac{1}{\sum p_{j, j=i,\cdots, |\mathcal{I}|}}$ and 
$p_j$ in the denominator is defined as the ratio between the number of devices in $\mathcal{S}_j$ and the number of participated devices in a global round, i.e., $\frac{|\mathcal{S}_j|}{|\mathcal{S}|}$.
\end{Lemma}

\begin{Lemma}
\label{lem2}
\textbf{(One round convergence)}. Under Assumptions \ref{ass1}-\ref{ass4} and Proposition 1, the divergence between the global model at the $(t+1)$-th global round and the optimal model satisfies $\mathbb{E} \left[ \Vert \mathbf{w}^{t+1} - \mathbf{w}^{*} \Vert^2 \right]   \leq  (1-\eta_t \mu \varepsilon) \mathbb{E} \left[\Vert \mathbf{w}^t - \mathbf{w}^* \Vert^2 \right] + \eta_t^2 (8 (\tau-1)^2G^2 + 2L\eta_t^2 ( |\mathcal{I}|\psi+ |\mathcal{S}| +\varepsilon ) \Lambda  + 2 \delta^2 \psi)$, where $L, \mu, \delta^2$, $\tau, G, \varepsilon$ are defined earlier and $\Lambda = \frac{1}{|\mathcal{S}|} \sum_{k \in \mathcal{S}} (F^* - F_k^*)$ measures the degree of non-i.i.d. in federated optimization. $F^* $ and $ F_k^*$ denote the optima of the global loss and local loss of device $k$, respectively.
\end{Lemma}

We direct readers to Appendix-\ref{Proof-Lemma1} and \ref{Proof-Lemma2} for the detailed proof of Lemmas \ref{lem1} and \ref{lem2}, respectively. Based on Lemmas \ref{lem1} and \ref{lem2}, the convergence rate of the proposed FedPMT is shown in the following Theorem \ref{thm1}, which is proven in Appendix-\ref{Proof-Theorem}.
\begin{customthm}{1}
\label{thm1}
Let Assumptions 1-5 hold and let $L, \mu, \delta_k, \tau, G, \varepsilon$ be as defined above. Choose the step size $\eta_t = \frac{2}{\mu \varepsilon (t+\lambda)}$, the convergence of federated learning with partial model training satisfies
\begin{align*}
\label{eq_themorem}
& \mathbb{E}\left[ F(\mathbf{w}^{T})  - F(\mathbf{w}^{*}) \right]  \leq \frac{1}{T+\lambda} \left( \frac{(\lambda +1 )\Gamma_1}{2} + \frac{2 \tilde{\Delta}}{ \mu^2}   \right), \tag{7}
\end{align*}
where $\lambda>0$, $ \tilde{\Delta} = (8 (\tau-1)^2G^2 + 2L ( |\mathcal{I}|\psi+|\mathcal{S}| +\varepsilon ) \Lambda + 2 \delta^2 \psi)/\varepsilon^2$,  and $\Gamma_{1} = \mathbb{E} \left[ \Vert \mathbf{w}^{1} - \mathbf{w}^{*} \Vert^2 \right]$ denotes the distance between the initial and optimal global models.
\end{customthm}

From Theorem 1, we observe that FedPMT has a convergence rate of $\mathcal{O}(1/T)$, which aligns with the convergence rate of FedAvg in \cite{fedprox, Li2020On}
(refer to Section \ref{resultC} for empirical verification). The difference between FedPMT and FedAvg lies in problem-related constant $\tilde \Delta$, essentially caused by information loss in partial model training. In addition, the bound in the right-hand side of (\ref{eq_themorem}) is related to model splitting (i.e., $\psi$), as analyzed in the following.

1) Given the initial global model $\mathbf{w}^1$, we have $\Gamma_1 = \Vert \mathbf{w}^1 - \mathbf{w}^* \Vert^2 \leq \frac{4}{\mu^2}G^2$ derived for a $\mu$-strongly convex global objective $F$ \cite{Li2020On}. Therefore, as shown in (\ref{eq_themorem}), the dominating term is $\mathcal{O}\left(\frac{ (L( |\mathcal{I}|\psi+|\mathcal{S}|+\varepsilon ) \Lambda + \delta^2 \psi + \lambda G^2 + \tau^2G^2 ) / \varepsilon^2}{ T \mu^2 }
\right)$, compared to the term $\mathcal{O}\left( \frac{L\Lambda + \delta^2 + \lambda G^2 + \tau^2G^2}{ T \mu^2 }
\right)$ in FedAvg. The results reveal that the loss gap between the global model $\mathbf{w}^{T}$ and optimal model $\mathbf{w}^{*}$ in FedPMT is more significant. This is because only a subset of participating devices update the whole model in local computation. Devices that update the partial model will lose information and contribute less to the global objective minimization. 

2) The loss gap in (\ref{eq_themorem}) is also related to the way to split the model, which determines how much the devices with partial model training can contribute to the global objective minimization. Notably, in order to shrink the loss gap between FedPMT and FedAvg, one needs to reduce $\psi$, i.e., $\sum_{i \in \mathcal{I}}  \frac{|\mathcal{I}||\mathcal{S}|}{\sum p_{j, j=i,\cdots, |\mathcal{I}|}}$, by enlarging the denominator of $\psi$. This demonstrates that the gap can be reduced with more devices updating more layers (i.e., a larger $\sum p_{j, j=i,\cdots, |\mathcal{I}|}$). On the contrary, if we assume devices $k\in \mathcal{S} \setminus \mathcal{S}_{|\mathcal{I}|} $ have the computational capability to do large computational tasks but they choose to do small tasks (e.g., updating the last layer of the model), this type of model splitting results in a smaller $\sum p_{j, j=i,\cdots, |\mathcal{I}|}$ and hence a larger gap. This analysis indicates that the partial model design should fully excavate the computation of the local devices in order to expedite the FL process. 

Even though FedPMT ends with a larger loss gap, with proper partial models being allocated to resource-constrained devices, FedPMT achieves a better trade-off in terms of completion time in FL.

\section{Numerical Results}

\label{IV}
In this section, we implement FedPMT across various tasks with different learning models and compare it with existing benchmarks FedAvg \cite{FedAvg} and a Dropout-based partial model training design, FedDrop \cite{dropout}. In particular, we use a Fully Connected Neural Network (FCNN)\footnote{FCNN model for MNIST task: $784 \times 400 $ Fully connected (Fc1) $\rightarrow 400 \times 300$ Fully connected (Fc2) $\rightarrow $ $\rm 300 \times 200$ Fully connected (Fc3) $\rightarrow 200\times100 $ Fully connected (Fc4) $\rightarrow 100\times10 $ Fully connected $\rightarrow$ Softmax. All Fully connected layers are mapped by ReLu activation.} and Convolutional Neural Network\footnote{CNN for MNIST task is constructed as below: $\rm 5 \times 5 \times 8$ Convolutional $\rightarrow 2 \times 2$ MaxPool $\rightarrow $ $\rm 5 \times 5 \times 16$ Convolutional $\rightarrow 2 \times 2$ MaxPool $\rightarrow 256\times 128 $ Fully connected $\rightarrow 128\times10 $ Fully connected $\rightarrow$ Softmax. 

CNN model for CIFAR10 task: $\rm 5 \times 5 \times 16$ Convolutional (Conv1) $\rightarrow 2 \times 2$ MaxPool $\rightarrow $ $\rm 5 \times 5 \times 32$ Convolutional (Conv2) $\rightarrow 2 \times 2$ MaxPool $\rightarrow 800\times 500 $ Fully connected (Fc1) $\rightarrow 500\times 300 $ Fully connected (Fc2) $\rightarrow 300\times10 $ Fully connected $\rightarrow$ Softmax. 

All Fully connected layers are mapped by ReLu activation.} (CNN) for MNIST and CIFAR-10 tasks, respectively. In the following Section, Section \ref{resultA}, we briefly describe the computational complexity analysis of the model learning process, including Forward Propagation (FP) and BP. 
Section \ref{resultB} describes the experiment setup. In Section \ref{resultC}, under the same computation setup,  we first compare FedPMT with FedDrop \cite{dropout} on MNIST dataset in terms of learning accuracy. Then, we compare FedPMT with FedAvg on the CIFAR10 dataset regarding task completion time for given target accuracies.

\begin{table*}[t]
\centering
\renewcommand{\arraystretch}{1.3}
\caption{Experiment setup to compare FedPMT and FedDrop\cite{dropout}. We set the same computational complexity (or keep a higher computation capability for FedDrop in cases when exact same complexity cannot be made) for FedPMT and FedDrop on each device to compare. \textit{rate} in FedDrop indicates that in order to keep the same computational complexity as FedPMT, FedDrop needs to keep $a$ percent of hidden layers' neurons, compared with the full model.}
\label{table1}
{\scriptsize
\begin{tabular}{c c c}
    \hline
  \multicolumn{3}{c}{ Computational complexity of FCNN-MNIST (local epoch $E=1$, batch size is 12)  } \\
Model Width ($|\mathcal{I}| = 4$)  & FedPMT complexity (ratio)   & FedDrop Complexity (dropout rate) \cite{dropout} \\
 \hline

Full - Fc1 (BP) - Fc2 (BP) - Fc3 (BP)   &  6473760 (42.3\%) & 6431556 ($\approx$ 54\%) \\
    Full - Fc1 (BP) -  Fc2 (BP)  &   7496160 (48.98\%) & 7579990 ($\approx$ 61\%) \\
   Full - Fc1 (BP) & 9779760 (63.9\%) & 9717454 ($\approx$ 73\%)\\ 
      FP+BP (Full) & 15305968 (100\%) & 100\% \\
    \hline
Model Width ($|\mathcal{I}| = 2$)  & &\\
 \hline
   Full - Fc1 (BP) & 9779760 (63.9\%) & 9717454 ($\approx$ 73\%)\\ 
      FP+BP (Full) & 15305968 (100\%) & 100\% \\
    \hline
    
 \multicolumn{3}{c}{  Computational complexity of CNN-MNIST (local epoch $E=1$, batch size is 12)  } \\
Model Width ($|\mathcal{I}| = 4$)  & PMT complexity (ratio)   & FedDrop Complexity (dropout rate) \\
 \hline

Full - Conv1(BP) - Conv2(BP) - Fc1 (BP)  &  745456 (40.8\%) & 1047049 (cap = 0.1) \\
    Full - Conv1(BP) - Conv2(BP) &   1188336 (65\%) & 1185944 ($\approx$ 0.26) \\
   Full - Conv1(BP) & 1597936 (87.4\%) & 1593950 ($\approx$ 0.73)\\ 
      FP+BP (Full) & 1828336 (100\%) & 1 \\
    \hline
    
    Model Width ($|\mathcal{I}| = 2$)  & &\\
 \hline
    Full - Conv1(BP) & 1597936 (87.4\%) & 1593950 ($\approx$ 0.73)\\ 
      FP+BP (Full) & 1828336 (100\%) & 1 \\    
    \hline   
      \multicolumn{3}{c}{ Computational complexity of CNN-CIFAR10 (local epoch $E=1$, batch size is 20)  } \\
Model Width ($|\mathcal{I}| = 5$)  & FedPMT complexity (ratio)  & FedDrop complexity (dropout rate) \cite{dropout}\\
 \hline

Full - Conv1(BP) - Conv2(BP) - Fc1 (BP) - Fc2 (BP)  &  12864200 (45.83\%) & 12885200 ($\approx$ 40\%) \\

Full - Conv1(BP) - Conv2(BP) - Fc1 (BP)  &  16077200 (57.27\%) & 16031980 ($\approx$ 54\%) \\
    Full - Conv1(BP) - Conv2(BP) &   24587200 (87.59\%) &  24677840 ($\approx$ 88\%) \\
   Full - Conv1(BP) & 26187200 (93.29\%) & 26069215 ($\approx$ 93\%)\\ 
      FP+BP (Full) & 28068800 (100\%) & 100\% \\
    \hline

Model Width ($|\mathcal{I}| = 2$)  & FedPMT complexity (ratio)  & FedDrop\\
 \hline
   Full - Conv1(BP) & 26187200 (93.29\%) & 26069215 ($\approx$ 93\%)\\ 
      FP+BP (Full) & 28068800 (100\%) & 100\% \\
    \hline
    
    \end{tabular}
}
\end{table*}

\subsection{Computational Complexity Analysis}
\label{resultA}
We consider the model in floating-point format (i.e., 32 bits for each parameter), and the operations in algorithms are floating-point operations. Following the similar analysis in \cite[Section IV-E]{PNS},
and supposing $n^x$ training samples in the calculation, we present the following complexity analysis.\\
\textbf{FP for FCNN}: 
\begin{itemize}
    \item The complexity of propagating from the input layer to the 2nd layer is represented as $O_{2,x} = W_{2,1}Z_{1,x}$, which has a complexity of $\mathcal{O}(n_2 \times n_1 \times n^x)$, where $Z, W, O$ represent input, weight parameter, and output of one layer, respectively. The subscript $\{2,1\}$ denotes the transition process between layers hereinafter, and $n_j$ is the number of neurons of the $j-$th layer.
    \item The activation function $Z_{2,x} = \bar f_{ac}(O_{2,x})$ has a complexity of $\mathcal{O}(n_2 \times n^x)$.
    \item The rest of the layers follow a similar analysis of the above steps.
\end{itemize}
    
\noindent \textbf{BP for FCNN}:  \\
For output layer (i.e., $o$) to the 4th hidden layer (Fc4), we 
\begin{itemize}
    \item Compute the error signal $e_{\{o,x\}}$ at the output layer as $e_{o,x} = \bar f_{ac}'(S_{o,x}) \circledast(Z_{o,x} - y_{o,x})$, where $Z_{o,x}$ is the raw output signal of the last layer, $\bar f_{ac}'$ is the inverse activation function, $y_{o,x}$ is the data label, and $\circledast$ represents element-wise multiplication.
    \item Compute the gradient $D_{o,4} = e_{o,x} \times Z_{x,4}$, where $Z_{x,4}$ is the transpose of $Z_{4,x}$.
    \item Update the weight on the 4th layer $W_{o,4} = W_{o,4} - \eta_t D_{o,4}$.
    
\end{itemize}

The complexity of the above operations is $\mathcal{O}(n_o\times n^x+n_o \times n^x+n_o \times n^x \times n_4+n_o \times n_4) $.

For the 4th hidden layer (Fc4) to 3rd hidden layer (Fc3), we have $e_{4,x} = \bar f_{ac}'(S_{4,x}) \circledast(W_{4,x} - e_{o,x})$, then $D_{4,3} = e_{4,x} \times Z_{x,3}$ and $W_{4,3} = W_{4,3} - \eta_t D_{4,3}$, where $ W_{4,3}$ is the transpose of $ W_{3,4}$. The complexity is $\mathcal{O}(n_4\times n^x+ n_4 \times n_o \times n^x+ n_4 \times n^x \times n_3+n_4 \times n_3)$. \\
The BP complexity of the rest of the layers of FCNN can be derived by a similar analogy.

\noindent \textbf{FP for CNN}: \\
The complexity of convolutional layers is found in \cite{he2015convolutional}, which is $\mathcal{O} (n_{l-1} \times s_l^2 \times n_l \times m_l^2 )$, where $l$ is the index of convolutional layer, $n_l$ indicates the number of filters in the $l$-th layer ($n_{l-1}$ is also known as the number of input channels in the $l$-th layer), $s_l$ is the spatial size of the filter, and $m_l$ is the spatial size of the output feature map, which is calculated as $m_l = (s_x - s_l +2\times padding )/stride +1$) and $s_x$ is the size of input.
\begin{itemize}
   \item Conv1: $n_{0}=3, n_1 = 16, s_1 = 5, m_1 = (32-5+2 \times 0)/1 +1 = 28$. Then,  using the max-pooling layer, the output feature size is $14\times14\times16$.
   \item Conv2: $n_{1}=16, n_2 = 32, s_2 = 5, m_2 = (14-5+2 \times 0)/1 +1 = 10$. Then,  using the max-pooling layer, the output feature size is $5 \times 5 \times 32$.
\end{itemize}

\noindent \textbf{BP for CNN}:  From \cite{he2015convolutional}, the complexity of the BP process for convolutional layers is roughly twice that of the FP process.

The FP and BP in the fully connected layer in CNN are the same as the cases in FCNN as discussed above.

The detailed computation is quantitively shown in Table \ref{table1}, where several 
training models with different model widths are provided. For example, $|\mathcal{I}|=4$ means four training model widths are available for the server (or devices) to choose. In Table I, \textit{FP+BP (Full)} represents devices with the full model, and \textit{Full - Fc1 (BP)} represents devices that \textit{do not} update the Fc1 layer. \textit{Full - Fc1 (BP) -Fc2 (BP)} represents devices that \textit{do not} update the Fc1 and Fc2 layers, and so on and so forth. The computational complexity of models with different model widths can be calculated according to the above discussion. In the meanwhile, to make a fair comparison, we set FedDrop \cite{dropout} with the same computational complexity as FedPMT. 

\begin{figure*}[t!]
\centerline{\includegraphics[scale=0.57]{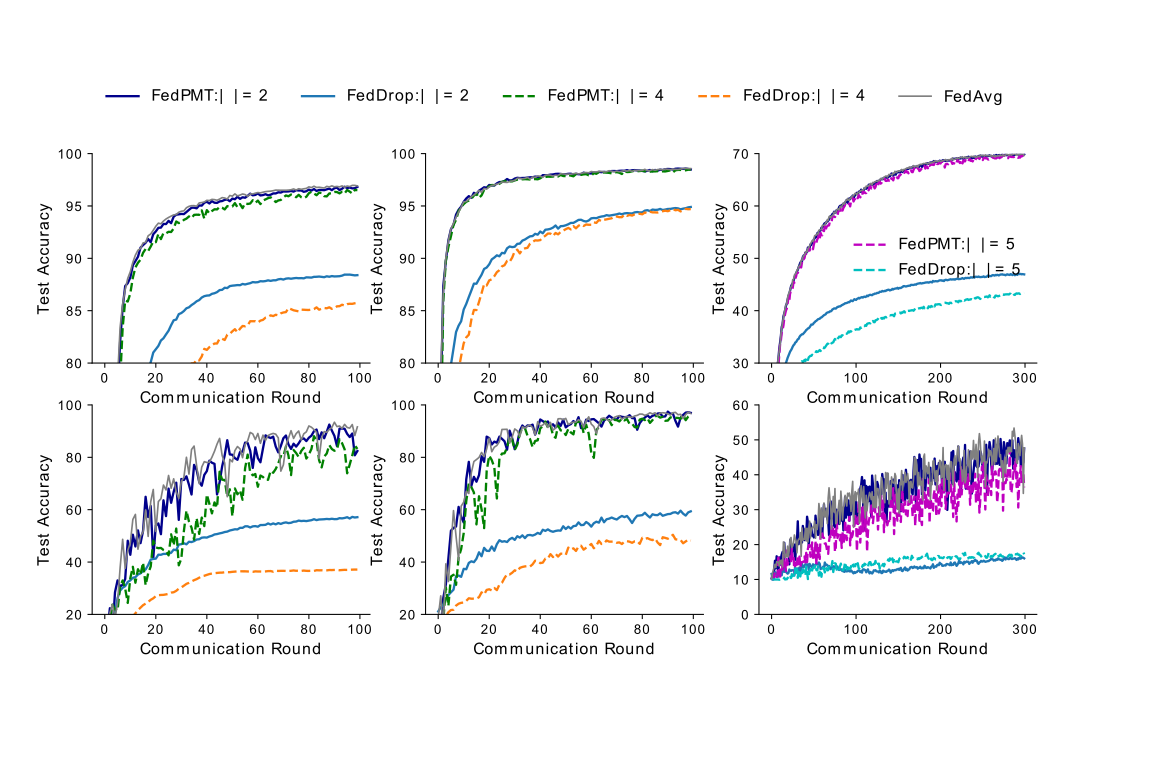}}
\caption{Test accuracy over communication rounds of FedPMT and FedDrop with different data heterogeneity and different computation levels in FL. From left to right, each column 
corresponds to the learning result on FCNN-MNIST, CNN-MNIST, and CNN-CIFAR10 tasks, respectively.
Upper and lower plots show the learning results for the i.i.d. and non-i.i.d. scenarios, respectively.}
\label{fig1}
\end{figure*}

\subsection{Experiment Setup}
\label{resultB}
\noindent \textbf{Data heterogeneity}: Two different data distribution settings are discussed, namely i.i.d. and non-i.i.d. settings. For the i.i.d. setting, data samples on each device are randomly selected from the training dataset. In the non-i.i.d. setting, data samples on each device belong to 2 different classes, which are randomly selected from 10 classes. The data samples on different devices form disjoint sets. 
We generate each setting and keep it fixed for different experiments to avoid randomness brought by training samples. Each experiment is executed with 10 random trails with fixed seeds in Pytorch.

For each experiment, a set of devices $\mathcal{S}$ is randomly selected in each global round from a set of candidate devices $\mathcal{K}$ with $|\mathcal{K}| = 100$. To better capture the impact of heterogeneous computation on FL learning performance, we assume that the number of selected devices with the same computation capabilities is evenly distributed among $|\mathcal{S}|$, e.g., $|\mathcal{S}_i| = 10 / |\mathcal{I}|, i \in \mathcal{I}$, in CIFAR10 experiments. The training setup is as follows, \\
\indent MNIST: $|\mathcal{D}_k| = 300, \eta_t = 0.01, E=1$, $|\mathcal{S}|=8$.\\
\indent  CIFAR10: $|\mathcal{D}_k| = 500, \eta_t = 0.05, E=1$, $|\mathcal{S}|=10$. \\
\noindent \textbf{FL Training time} (CNN-CIFAR10 task)\\
$\kappa$ setting: Since FedPMT targets reducing the training time for computation heterogeneous FL, we set five different computation levels, $0.2\Psi, 0.25\Psi, 0.33\Psi, 0.5\Psi, 1\Psi$,  where $1\Psi$ represents the maximum computation capability for a set of participating devices. For example, suppose a device with $1 \Psi$ can complete the local training in 10 seconds (i.e., $ T_{cmp} = \frac{c_k \cdot |\mathcal{D}_k| \cdot E}{\kappa_k} $ = 10), devices with $0.2\Psi$ takes 50 seconds to finish the same task. \\
$c$ setting: The computation time in FedPMT is analyzed as follows: Since the devices with model width smaller than full model width only need to update part of the whole model, which makes $c_k$ smaller. For the case with five different model widths $|\mathcal{I}|=5$ (see Table \ref{table1}), the training models with the complexity ratio 0.46, 0.58, 0.88, 0.94, and 1 will be assigned to devices with $0.2\Psi, 0.25\Psi, 0.33\Psi, 0.5\Psi$, and $1\Psi$, respectively. Therefore, the training time consumption is $0.46 \times 50 s, 0.58\times40s, 0.88\times30s, 0.94\times20s$, and $1\times10s$, respectively (assuming $1\Psi$ can complete the local training of a full model in 10 seconds).

\subsection{Empirical Results}
\label{resultC}
We first compared FedPMT with a dropout-based algorithm, FedDrop \cite{dropout}, in scenarios where different participating devices have different computational levels, reflected by different $|\mathcal{I}|$. Given the above setup, FedPMT generates different partial models for computation-heterogeneous devices, where devices with small computation capacity will restrict gradient from back-propagating to more shallow layers. While FedDrop\cite{dropout} creates different partial models by removing varying numbers of neurons in hidden layers to match devices' computation capabilities.

As shown in Fig. \ref{fig1}, FedPMT
outperforms FedDrop across different computation heterogeneity and data distribution settings. Both FedPMT and FedDrop perform better in the cases with model width $|\mathcal{I}|=2$, compared to cases with model width $|\mathcal{I}|=4$. The FL global model converges faster since devices' computation capabilities are generally higher in $|\mathcal{I}|=2$. With limited computation power on devices, FedDrop randomly removes neurons in hidden layers, making model capacity small. While devices in FedPMT sacrifice shallow layers and prioritize the most crucial layers, thus ensuring a better performance than FedDrop. Those non-prioritized layers can still be updated in the model aggregation. The inaccuracy in shallow layers impacts model performance less than that in deep layers (as seen in FedDrop). This observation is more evident with non-i.i.d. data. FedPMT with model width $|\mathcal{I}| = 4$ achieves more than 90\% accuracy, while FedDrop barely works with an accuracy lower than 60\%. This is because data samples share common features in the non-i.i.d. case, and each local classifier (the last layer) is more sensitive to different data distributions. Given limited computation power, we need to prioritize the crucial layers (near classifier) instead of evenly reducing the number of neurons in hidden layers as done in FedDrop. In addition, regardless of the learning completion time, FedAvg's learning result is provided as an upper bound for different tasks. FedAvg assumes homogeneous models across local devices and does not consider devices' heterogeneous computation capabilities. As can be seen in Fig. \ref{fig1}, FedPMT with smaller model widths (e.g., $|\mathcal{I}| = 2$) achieves similar learning results as FedAvg for all i.i.d. and non-i.i.d. cases. Among CNN model-related tasks, FedPMT achieves very competitive results even for more computation heterogeneous scenarios ($|\mathcal{I}| = 4$ or $5$), although fluctuations in the learning process are observed in non-i.i.d. scenarios, leaving the performance margin to FedAvg negligible, compared to extra computational complexity in FedAvg. For the FCNN-MNIST task with more heterogeneous computation, there is a larger performance gap between FedPMT and FedAvg. However, the proposed design still outperforms FedDrop with a prominent performance gap.

\begin{table}[h]
\centering
\renewcommand{\arraystretch}{1.3}
\caption{Learning time comparison between different FL designs}
\label{table2}
{\scriptsize
\begin{tabular}{c |  c c c c}
    \hline

  & \multicolumn{2}{c}{  constraint (26.5 seconds)} & \multicolumn{2}{c}{ without constraint}  \\

Accuracy  & FedPMT  & FedAvg & FedPMT  & FedAvg \\
  & \multicolumn{4}{c}{ i.i.d.}  \\
 \hline
  50\%  &  1064.8  & 1121.8  & 1029.6  & 1950  \\
   55\% & 1584 & 1732 & 1601.6 & 2800 \\ 
    60\% &   2270.4   &  2464.5  & 2217.6  & 4266.6  \\
    \hline
& \multicolumn{4}{c}{non-i.i.d.} 
   \\ \hline
40\% & 4136 & 4507 &  4174 & 4644 \\ 
 45\%  & 6157 & 6604.4 & 6218.7 & 6577.7\\  
 50\%  & 8251 & 8771.5 &   9234.1 & 10361 \\   \hline
 \end{tabular}
}
\end{table}
Next, we compare FedPMT and FedAvg on completion time in FL. We consider two different cases: 1) with a constraint (26.5 seconds)\footnote{This constraint is set as the model training time spent by the devices with the longest completion time in FedPMT, as calculated in Section \ref{resultB}.}, the computation time constraint is set in each global round. Beyond this time stamp, the server aggregates the received models (without waiting for the rest) and moves to the next global round. 2) \textit{without constraint} means that the server aggregates models after receiving all local models in each global round. 

With a time constraint in each round, more devices in FedPMT can contribute to the global model aggregation, even though models from devices with limited computation capabilities are not completely updated. FedPMT is more effective in the non-i.i.d. case, where aggregating more local models results in faster convergence, as also observed in \cite{FedAvg, 8664630}. If no constraint is set in each global round, FedPMT can complete the learning task in almost half the time, compared to FedAvg (2217.6 seconds vs. 4266.6 seconds at 60\% accuracy). Although FedAvg obtains more accurate local models in each round, it is inefficient in terms of completion time. In contrast, FedPMT achieves a better trade-off between model accuracy and completion time.

\section{Conclusion}
\label{V}
In this paper, we have presented our model-heterogeneous FL design, FedPMT, which enables computation-constrained devices to participate in federated learning and contribute to the global model. As a partial model training strategy, FedPMT achieves sub-model training from the backpropagation perspective. Unlike Dropout-based partial model training that randomly removes neurons in hidden layers, FedPMT allows all participating devices to prioritize the most crucial parts (deep layers) of the global model, ensuring a relatively large model capacity. We have analyzed the convergence rate of FedPMT, which shows a similar convergence property as FedAvg, with a slightly larger sub-optimality gap factored with a model splitting-related constant. Our experimental results show that FedPMT consistently outperforms the state-of-the-art Dropout-based algorithm, FedDrop. Meanwhile, FedPMT reaches the learning target with a shorter completion time and achieves a better trade-off between the learning accuracy and FL training time compared to the widely adopted model-homogeneous benchmark, FedAvg.

{\appendix
\subsection{Proof of Lemma 1}
\label{Proof-Lemma1}
For the ease of analysis, the gradient $\nabla F_k(\mathbf{w}_{k}^{t}, \xi_k)$ and $\nabla F_k(\mathbf{w}_{k}^{t})$ are represented by ${g}_{k,i}^{t}$ and $\bar{g}_{k,i}^{t}$ in the following proof, where the subscript $i$ in ${g}_{k,i}^{t}$ indicates that the device $k$ belongs to set $\mathcal{S}_i$.

From the definition of $\nabla F(\mathbf{w}^{t})$, we have
\begin{align*}
\label{a1}
& \Vert \nabla F(\mathbf{w}^{t}) - \nabla \bar F(\mathbf{w}^{t}) \Vert^2 \\
 =  & \,  \Vert \sum_{i \in \mathcal{I}}  \frac{1}{|\cup \mathcal{S}_{j, j=i,\cdots, |\mathcal{I}|} |}  \sum_{k \in \cup \mathcal{S}_j}  ({g}_{k,i}^{t} - \bar{g}_{k,i}^{t}) \circ \Upsilon_k^{|\mathcal{I}|-i+1} \Vert^2 \\ 
\stackrel{1} \leq & \,  |\mathcal{I}| \cdot \sum_{i \in \mathcal{I}}  \frac{1}{|\cup \mathcal{S}_{j, j=i,\cdots, |\mathcal{I}|} |}  \Vert  \sum_{k \in \cup \mathcal{S}_j}  ( {g}_{k,i}^{t} - \bar{g}_{k,i}^{t}) \circ \Upsilon_k^{|\mathcal{I}|-i+1} \Vert^2\\ 
 \stackrel{2} \leq  & \,  \sum_{i \in \mathcal{I}}  \frac{|\mathcal{I}| |\mathcal{S}|}{|\cup \mathcal{S}_{j, j=i,\cdots, |\mathcal{I}|} |}  \sum_{k \in \cup \mathcal{S}_j}  \Vert ( {g}_{k,i}^{t} - \bar{g}_{k,i}^{t}) \circ \Upsilon_k^{|\mathcal{I}|-i+1} \Vert^2\\ 
 \stackrel{3} \leq & \hquad  \sum_{i \in \mathcal{I}}  \frac{ |\mathcal{I}| |\mathcal{S}| }{|\cup \mathcal{S}_{j, j=i,\cdots, |\mathcal{I}|} |}  \sum_{k \in \cup \mathcal{S}_j}  \Vert {g}_{k,i}^{t} - \bar{g}_{k,i}^{t} \Vert^2, \tag{B1}
\end{align*}
where inequality 1 holds by Cauchy-Schwartz inequality, inequality 2 holds by Cauchy-Schwartz inequality and $\cup \mathcal{S}_{j, j=i,\cdots, |\mathcal{I}|} \subset \mathcal{S}, \forall i \in \mathcal{I}$, and inequality 3 holds because the norm of partial gradient is smaller than the norm of full gradient, i.e., 
$\Vert ( {g}_{k,i}^{t} - \bar{g}_{k,i}^{t}) \circ \Upsilon_k^{|\mathcal{I}|-i+1} \Vert^2 < \Vert ( {g}_{k,i}^{t} - \bar{g}_{k,i}^{t})\Vert^2$ for all model width $i\in \mathcal{I}$.

Taking the expectation on both sides of (\ref{a1}), we have
\begin{align*}
\label{a3}
& \mathbb{E} \left[ \Vert \nabla F(\mathbf{w}^{t}) - \nabla \bar 
 F(\mathbf{w}^{t}) \Vert^2 \right] \\ 
 \leq & \,   \mathbb{E} \left[ \sum_{i \in \mathcal{I}}  \frac{ |\mathcal{I}| |\mathcal{S}| }{|\cup \mathcal{S}_{j, j=i,\cdots, |\mathcal{I}|} |}  \sum_{k \in \cup \mathcal{S}_j}  \Vert {g}_{k,i}^{t} - \bar{g}_{k,i}^{t}  \Vert^2
\right] \\
 =  & \,  \sum_{i \in \mathcal{I}}  \frac{ |\mathcal{I}| |\mathcal{S}| }{|\cup \mathcal{S}_{j, j=i,\cdots, |\mathcal{I}|} |}  \sum_{k \in \cup \mathcal{S}_j} \mathbb{E} \left[ \Vert {g}_{k,i}^{t} - \bar{g}_{k,i}^{t}  \Vert^2 \right] \\
 \stackrel{4}\leq  & \hquad  \sum_{i \in \mathcal{I}}  \frac{ |\mathcal{I}| |\mathcal{S}| }{|\cup \mathcal{S}_{j, j=i,\cdots, |\mathcal{I}|} |}  \sum_{k \in \mathcal{S}} \mathbb{E} \left[ \Vert {g}_{k,i}^{t} - \bar{g}_{k,i}^{t}  \Vert^2 \right] \\
 \stackrel{5} = & \, \sum_{i \in \mathcal{I}}  \frac{|\mathcal{I}| |\mathcal{S}| }{|\mathcal{S}| \cdot \sum_{j=i}^{|\mathcal{I}|} p_j}  \sum_{k \in \mathcal{S}} \mathbb{E} \left[ \Vert {g}_{k,i}^{t} - \bar{g}_{k,i}^{t}  \Vert^2 \right] \\
 = & \, \frac{1}{|\mathcal{S}| } \sum_{i \in \mathcal{I}}  \frac{|\mathcal{I}| |\mathcal{S}| }{ \sum_{j=i}^{|\mathcal{I}|} p_j}  \sum_{k \in \mathcal{S}} \delta_k^2\\
\leq & \,  2\delta^2 \sum_{i \in \mathcal{I}}  \frac{|\mathcal{I}||\mathcal{S}| }{\sum_{j=i}^{|\mathcal{I}|} p_j}, \tag{B2}
\end{align*}
where inequality 4 holds by $\cup \mathcal{S}_{j, j=i,\cdots, |\mathcal{I}|} \subset \mathcal{S}, \forall i \in \mathcal{I}$, and  
$p_j$ in the denominator in equality 5 is defined as the ratio between the number of devices in $\mathcal{S}_j$ and the number of participated devices in a global round, i.e., $\frac{|\mathcal{S}_j|}{|\mathcal{S}|}$.

\subsection{Proof of Lemma 2}
\label{Proof-Lemma2}
By the definition of $\mathbb{E}[ \nabla F(\mathbf{w}^{t})] = \nabla \bar F(\mathbf{w}^{t}) $, we have
\begin{align*}
\label{B1}
& \Vert \mathbf{w}^{t+1} - \mathbf{w}^{*} \Vert^2 \\
=  & \hquad \Vert \mathbf{w}^{t}- \eta_t \nabla F(\mathbf{w}^{t}) - \mathbf{w}^{*} - \eta_t \nabla \bar F(\mathbf{w}^{t}) + \eta_t \nabla \bar F(\mathbf{w}^{t}) \Vert^2 \\
=  & \hquad \underbrace{\Vert \mathbf{w}^{t}- \eta_t \nabla \bar F(\mathbf{w}^{t}) - \mathbf{w}^{*} \Vert^2}_{\mathfrak{C}_1} + \eta_t^2 \underbrace{\Vert \nabla F(\mathbf{w}^{t}) - \nabla \bar F(\mathbf{w}^{t})\Vert^2}_{\mathfrak{C}_2}  \\ &  + \underbrace{ 2\eta_t \langle \mathbf{w}^{t}- \eta_t \nabla \bar F(\mathbf{w}^{t}) - \mathbf{w}^{*}, \nabla \bar F(\mathbf{w}^{t}) - \nabla F(\mathbf{w}^{t}) \rangle}_{\mathfrak{C}_3}\\
=  & \hquad \Vert \mathbf{w}^{t} - \mathbf{w}^{*} \Vert^2  + \eta_t^2 \Vert \nabla \bar F(\mathbf{w}^{t}) \Vert^2 - 2 \eta_t \langle \mathbf{w}^{t} - \mathbf{w}^{*},  \nabla \bar F(\mathbf{w}^{t}) \rangle \\ & + \eta_t^2 \mathfrak{C}_2 + \mathfrak{C}_3. \tag{C1}
\end{align*}

$\bullet$ \noindent Bounding term $\eta_t^2 \Vert \nabla \bar F(\mathbf{w}^{t}) \Vert^2$ 

By the definition of $\nabla \bar F(\mathbf{w}^{t})$, we have 
\begin{align*}
\label{B3}
& \Vert \nabla \bar F(\mathbf{w}^{t}) \Vert^2 \\
= & \, \Vert \sum_{i \in \mathcal{I}}  \frac{1}{|\cup \mathcal{S}_{j, j=i,\cdots, |\mathcal{I}|} |}  \sum_{k \in \cup \mathcal{S}_j} \nabla F_k(\mathbf{w}_{k}^{t}) \circ \Upsilon_k^{|\mathcal{I}|-i+1 } \Vert^2 \\
\stackrel{1,2,3} \leq & \,  \sum_{i \in \mathcal{I}}  \frac{ |\mathcal{I}| |\mathcal{S}| }{|\cup \mathcal{S}_{j, j=i,\cdots, |\mathcal{I}|} |}  \sum_{k \in \cup \mathcal{S}_j} \Vert \nabla F_k(\mathbf{w}_{k}^{t})\Vert^2 \\
\stackrel{4,5} \leq  & \, \frac{1}{|\mathcal{S}|} \sum_{i \in \mathcal{I}}  \frac{|\mathcal{I}| |\mathcal{S}|}{\sum_{j=i}^{|\mathcal{I}|} p_j}  \sum_{k \in \mathcal{S}} \Vert  \nabla F_k(\mathbf{w}_{k}^{t})\Vert^2. \tag{C2}
\end{align*}

$\Vert \nabla F_k(\mathbf{w}_{k}^{t}) \Vert^2$ in equation (\ref{B3}) is bounded as follows. Given any models $\mathbf{w}_k^t$ and $\mathbf{w}_k^{\prime}$ satisfying Assumption \ref{ass2}, we have $F_k( \mathbf{w}_k^{\prime}) - F_k( \mathbf{w}_k^t) -  (\mathbf{w}_k^{\prime} - \mathbf{w}_k^t)^\top \nabla F_k(\mathbf{w}_{k}^{t}) \leq  \frac{L}{2} \Vert \mathbf{w}_k^{\prime} - \mathbf{w}_k^t \Vert^2 $. By defining  $\mathbf{w}_k^{\prime} = \mathbf{w}_k^t - \frac{1}{L} \nabla F_k(\mathbf{w}_k^t)$, we have $ F_k(\mathbf{w}_k^{\prime}) - F_k(\mathbf{w}_k^t) \leq - \frac{1}{L} (\nabla F_k(\mathbf{w}_k^t))^\top \nabla F_k(\mathbf{w}_k^t) + \frac{L}{2} \cdot \frac{1}{L^2} \Vert \nabla F_k(\mathbf{w}_k^t) \Vert^2 \leq -\frac{1}{2L} \Vert \nabla F_k(\mathbf{w}_k^t) \Vert^2$. Taking the minimal loss $F_k^*$ on device $k$, we have
\begin{align*}
\label{B4}
\Vert \nabla F_k(\mathbf{w}_{k}^{t}) \Vert^2  \leq 2L (F_k(\mathbf{w}_k^t) - F_k(\mathbf{w}_k^{\prime})) \leq 2L (F_k(\mathbf{w}_k^t) - F_k^*). \tag{C3}
\end{align*}

We use $\psi$ to denote the constant  $\sum_{i \in \mathcal{I}}  \frac{|\mathcal{I}||\mathcal{S}| }{\sum_{j=i}^{|\mathcal{I}|} p_j}$ hereinafter. As such, $\eta_t^2 \Vert \nabla \bar F(\mathbf{w}^{t}) \Vert^2$ is bounded by combing (\ref{B4}) and (\ref{B3}), and we have
\begin{align*}
\label{B5}
\eta_t^2 \Vert \nabla \bar F(\mathbf{w}^{t}) \Vert^2 
& \leq 2 |\mathcal{I}| L \eta_t^2 \psi \frac{1}{|\mathcal{S}|}  \sum_{k \in \mathcal{S}}
(F_k(\mathbf{w}_k^t) - F_k^*) . \tag{C4}
\end{align*}

$\bullet$ \noindent Bounding term $- 2 \eta_t \langle \mathbf{w}^{t} - \mathbf{w}^{*},  \nabla \bar F(\mathbf{w}^{t}) \rangle$. 

Again, by the definition of $\nabla \bar F(\mathbf{w}^{t})$ and (\ref{eq7}), we have 
\begin{align*}
\label{B6}
& -2 \eta_t \langle \mathbf{w}^{t} - \mathbf{w}^{*}, \nabla \bar F(\mathbf{w}^{t}) \rangle  \\
= & -2 \eta_t \langle \mathbf{w}^{t} - \mathbf{w}^{*}, 
\sum_{k \in \mathcal{S}} A_k \circ \nabla F_k(\mathbf{w}_k^{t}) \rangle  \\
= & \underbrace{-2 \eta_t \sum_{k \in \mathcal{S}}  \langle \mathbf{w}^{t} - \mathbf{w}_k^{t}, \tilde{\tilde{\nabla F_k}}(\mathbf{w}_k^{t}) \rangle }_{\mathfrak{C}_{4.1}}    \underbrace{-2 \eta_t  \sum_{k \in \mathcal{S}}  \langle \mathbf{w}_k^{t} -\mathbf{w}^{*}, \tilde{\tilde{\nabla F_k}}(\mathbf{w}_k^{t})  \rangle}_{\mathfrak{C}_{4.2}}. 
\tag{C5}
\end{align*}
where $\tilde{\tilde{\nabla F_k}}(\mathbf{w}_k^{t})  = A_k \circ \nabla F_k(\mathbf{w}_k^{t}) $ is the result of local ground-truth gradient after layer-wise multiplication with weight $A_k$.

Each term in $\mathfrak{C}_{4.1}$ is bounded as follows: By Cauchy-Schwarz inequality, AM-GM inequality, we have the first inequality hold in (\ref{C6}). The last inequality in (\ref{C6}) is achieved since $\Vert \tilde{\tilde{\nabla F_k}}(\mathbf{w}_k^{t}) \Vert^2 = \Vert A_k \circ \nabla F_k(\mathbf{w}_k^{t})\Vert^2 < \Vert \nabla F_k(\mathbf{w}_k^{t}) \Vert^2$.
\begin{align*}
\label{C6}
& -2 \eta_t  \langle \mathbf{w}^{t} - \mathbf{w}_k^{t}, \tilde{\tilde{\nabla F_k}}(\mathbf{w}_k^{t})   \rangle \\ \leq & \, \eta_t(\frac{1}{\eta_t} \Vert \mathbf{w}^{t} - \mathbf{w}_k^{t} \Vert^2 + \eta_t \Vert \tilde{\tilde{\nabla F_k}}(\mathbf{w}_k^{t})  \Vert^2) 
\\  \leq & \, \Vert \mathbf{w}^{t} - \mathbf{w}_k^{t} \Vert^2 + \eta_t^2  \Vert \nabla F_k(\mathbf{w}_{k}^{t}) \Vert^2.
\tag{C6} 
\end{align*}

By Assumption 1 and Proposition 1, each term in $\mathfrak{C}_{4.2}$ is bounded as
\begin{align*}
& -2 \eta_t  \langle \mathbf{w}_k^{t}  - \mathbf{w}^{*}, \tilde{\tilde{\nabla F_k}}(\mathbf{w}_k^{t})  \rangle 
\\ \leq & \,  2 \eta_t \varepsilon  \frac{1}{|\mathcal{S}|} \left( (-(F_k(\mathbf{w}_k^{t}) - F_k(\mathbf{w}^{*}) ) - \frac{\mu}{2} \Vert \mathbf{w}_k^{t} - \mathbf{w}^* \Vert^2) \right). \tag{C7}
\end{align*}


Based on the above intermediate results, $-2 \eta_t \langle \mathbf{w}^{t} - \mathbf{w}^{*}, \nabla \bar F(\mathbf{w}^{t}) \rangle $ is bounded as 
\begin{align*}
\label{B9}
& -2 \eta_t \langle \mathbf{w}^{t} - \mathbf{w}^{*}, \nabla \bar F(\mathbf{w}^{t}) \rangle \\ 
\leq &  \sum_{k\in \mathcal{S}}  (\Vert \mathbf{w}^{t} - \mathbf{w}_k^{t} \Vert^2 + \eta_t^2 \Vert \nabla F_k(\mathbf{w}_{k}^{t}) \Vert^2 \\ & -2 \eta_t \varepsilon \frac{1}{|\mathcal{S}|} (F_k(\mathbf{w}_k^{t}) - F_k(\mathbf{w}^{*}) ) - \mu \eta_t \varepsilon \frac{1}{|\mathcal{S}|} \Vert \mathbf{w}_k^{t} - \mathbf{w}^* \Vert^2 ) \\
= & - \mu \eta_t \varepsilon\Vert \mathbf{w}^{t} - \mathbf{w}^* \Vert^2 + \sum_{k\in \mathcal{S}} (\Vert \mathbf{w}^{t} - \mathbf{w}_k^{t} \Vert^2 + \eta_t^2 \frac{|\mathcal{S}|}{|\mathcal{S}|}  \Vert \nabla F_k(\mathbf{w}_{k}^{t}) \Vert^2 \\ & -2 \eta_t \varepsilon \frac{1}{|\mathcal{S}|}  (F_k(\mathbf{w}_k^{t}) - F_k(\mathbf{w}^{*}) ) 
. \tag{C8}
\end{align*}

Inserting (\ref{B4}), (\ref{B5}), and (\ref{B9})  to (\ref{B1}), we have 
\begin{align*}
\label{B10}
& \Vert \mathbf{w}^{t+1} - \mathbf{w}^{*} \Vert^2 
\\ \leq  & (1-\eta_t \mu )\Vert \mathbf{w}^{t} - \mathbf{w}^{*} \Vert^2 + \sum_{k\in \mathcal{S}} \Vert \mathbf{w}^{t} - \mathbf{w}_k^{t} \Vert^2 + \eta_t^2 \mathfrak{C}_1 + \mathfrak{C}_2 \\ & + (2|\mathcal{I}| L \eta_t^2 \psi + 2 L \eta_t^2 |\mathcal{S}| ) \frac{1}{|\mathcal{S}|}  \sum_{k \in \mathcal{S}}
(F_k(\mathbf{w}_k^t) - F_k^*) \\ & 
-2 \eta_t \varepsilon \frac{1}{|\mathcal{S}|} \sum_{k\in \mathcal{S}} (F_k(\mathbf{w}_k^{t}) - F_k(\mathbf{w}^{*}) ) . \tag{C9}
\end{align*}
where the summation of the last two terms in the right-hand side of (\ref{B10}) is labeled as $\mathfrak{C}_5$ in the following. 

Defining $\gamma_t = 2 \eta_t(\varepsilon - \eta_t L( |\mathcal{I}| \psi +|\mathcal{S}|) )$. In addition, we have $\eta_t \leq \frac{\varepsilon}{2L (|\mathcal{I}| \psi + |\mathcal{S}|})$ and $ \eta_t \varepsilon \leq \gamma_t \leq 2\eta_t \varepsilon$. $\mathfrak{C}_{5}$ is transformed as
\begin{align*}
\label{B11}
& \mathfrak{C}_{5} \\ 
= & -\gamma_t \frac{1}{|\mathcal{S}|}  \sum_{k \in \mathcal{S}} (F_k(\mathbf{w}_k^t) - F_k^* ) + 2\eta_t \varepsilon \frac{1}{|\mathcal{S}|}  \sum_{k \in \mathcal{S}} (F_k(\mathbf{w}_k^t) - F_k^* ) \\ & \quad -2 \eta_t \varepsilon \frac{1}{|\mathcal{S}|} \sum_{k\in \mathcal{S}} (F_k(\mathbf{w}_k^{t}) - F_k(\mathbf{w}^{*}) )  \\
=  & -\gamma_t \frac{1}{|\mathcal{S}|}  \sum_{k \in \mathcal{S}} (F_k(\mathbf{w}_k^t) - F_k^* + F^* - F^*) \\& \quad  + 2\eta_t \varepsilon \frac{1}{|\mathcal{S}|}  \sum_{k \in \mathcal{S}} (F_k(\mathbf{w}^{*})- F_k^* )
\\ = & -\gamma_t \frac{1}{|\mathcal{S}|}  \sum_{k \in \mathcal{S}} (F_k(\mathbf{w}_k^t) - F^*)  + (2\eta_t \varepsilon -\gamma_t ) \frac{1}{|\mathcal{S}|}  \sum_{k \in \mathcal{S}} (F^{*}- F_k^* )
\\ = &  -\gamma_t \frac{1}{|\mathcal{S}|}  \sum_{k \in \mathcal{S}} (F_k(\mathbf{w}_k^t) - F^*) \\& \quad + 2L\eta_t^2 ( |\mathcal{I}| \psi + |\mathcal{S}|) \frac{1}{|\mathcal{S}|}  \sum_{k \in \mathcal{S}} (F^{*}- F_k^* )
\\ = &  -\gamma_t \underbrace{\frac{1}{|\mathcal{S}|}  \sum_{k \in \mathcal{S}} (F_k(\mathbf{w}_k^t) - F^*) }_{\mathfrak{C}_{5.1}} + 2L\eta_t^2 ( |\mathcal{I}| \psi + |\mathcal{S}|)   \Lambda,
\tag{C10}
\end{align*}
where $\Lambda = \frac{1}{|\mathcal{S}|} \sum_{k \in \mathcal{S}} (F^* - F_k^*)$ measures the degree of non-i.i.d. in federated optimization.
$F^*$, $F_k^*$, and $F_k(\mathbf{w}^*)$ represent the optional global loss, the optional local loss on device $k$, and the local loss on device $k$ with optimal model $\mathbf{w}^*$, respectively.

To bound $\mathfrak{C}_{5.1}$, we have 
\begin{align*}
\label{B12}
& \frac{1}{|\mathcal{S}|} \sum_{k \in \mathcal{S}} (F_k(\mathbf{w}_k^t) - F^*) \\
= & \hquad \frac{1}{|\mathcal{S}|} \sum_{k \in \mathcal{S}} (F_k(\mathbf{w}_k^t) -  F_k(\mathbf{w}^t)) + \frac{1}{|\mathcal{S}|} \sum_{k \in \mathcal{S}} (F_k(\mathbf{w}^t) - F^*)  \\
\geq & \hquad  \frac{1}{|\mathcal{S}|} \sum_{k \in \mathcal{S}} (\langle \nabla F_k(\mathbf{w}^t), \mathbf{w}_k^t - \mathbf{w}^t \rangle  + F(\mathbf{w}^t) - F^*) \\
\stackrel{6} \geq &  -\frac{1}{2} \frac{1}{|\mathcal{S}|} \sum_{k \in \mathcal{S}} ( \eta_t \Vert \nabla \bar F_k(\mathbf{w}^t) \Vert^2 + \frac{1}{\eta_t } \Vert \mathbf{w}_k^t - \mathbf{w}^t \Vert^2)  \\ & + \frac{1}{|\mathcal{S}|} \sum_{k \in \mathcal{S}} (F(\mathbf{w}^t) - F^*)  \ \\
 \geq  & \, - \frac{1}{|\mathcal{S}|} \sum_{k \in \mathcal{S}} [ \eta_t L (F_k(\mathbf{w}^t) - F_k^*) + \frac{1}{2\eta_t} \Vert \mathbf{w}_k^t - \mathbf{w}^t \Vert^2  \\&  + F(\mathbf{w}^t) - F^* ],
\tag{C11}
\end{align*}
where the first inequality results from the convexity of local loss $F_k$, inequality 6 is held by AM-GM inequality, and the last inequality is achieved by (\ref{B4}).

By combing (\ref{B12}) and (\ref{B11}), $\mathfrak{C}_{5}$ is bounded as
\begin{align*}
\label{B13}
& \mathfrak{C}_{5} \\ 
\leq & \, \gamma_t \frac{1}{|\mathcal{S}|} \sum_{k \in \mathcal{S}}[\eta_t L (F_k(\mathbf{w}^t) - F_k^*) + \frac{1}{2\eta_t} \Vert \mathbf{w}_k^t - \mathbf{w}^t \Vert^2 ]\\&  - \gamma_t (F(\mathbf{w}^t) - F^*) + 2L\eta_t^2 ( |\mathcal{I}| \psi + |\mathcal{S}|)   \Lambda \\
= &  \, \gamma_t  \frac{1}{|\mathcal{S}|} \sum_{k \in \mathcal{S}} [  \eta_tL (F_k(\mathbf{w}^t) - F^* + F^* - F_k^*)  + \frac{1}{2\eta_t} \Vert \mathbf{w}_k^t - \mathbf{w}^t \Vert^2 \vphantom{\eta_tL} ]  \\ & - \gamma_t  (F(\mathbf{w}^t) - F^*) + 2L\eta_t^2 ( |\mathcal{I}| \psi + |\mathcal{S}|) \Lambda \\
= & \, \gamma_t  (\eta_t L -1) \frac{1}{|\mathcal{S}|} \sum_{k \in \mathcal{S}} (F_k(\mathbf{w}^t) - F^*) + \frac{\gamma_t }{2\eta_t} \frac{1}{|\mathcal{S}|} \sum_{k \in \mathcal{S}} \Vert \mathbf{w}_k^t  - \mathbf{w}^t\Vert^2 \\ & + [\gamma_t \eta_t L +  2L\eta_t^2 ( |\mathcal{I}| \psi + |\mathcal{S}|) ]\Lambda \\
\leq & \hquad \frac{1}{|\mathcal{S}|} \sum_{k \in \mathcal{S}} \Vert \mathbf{w}_k^t - \mathbf{w}^t \Vert^2 +  2L\eta_t^2 ( |\mathcal{I}|\psi+|\mathcal{S}|+\varepsilon ) \Lambda ,
\tag{C12}
\end{align*}
where the last inequality achieves because: 1) We have $\gamma_t>0$ since $ \eta_t \varepsilon \leq \gamma_t \leq 2\eta_t \varepsilon$, and $\eta_t L -1 =  \frac{ \varepsilon}{2(|\mathcal{I}| \psi +|\mathcal{S}|)} -1 \leq 0$, so that $\gamma_t  (\eta_t L -1) \frac{1}{|\mathcal{S}|} \sum_{k \in \mathcal{S}} (F_k(\mathbf{w}^t) - F^*) \leq 0$. 2) Since $ \eta_t \varepsilon \leq \gamma_t \leq 2\eta_t \varepsilon$, we have $\frac{\gamma_t }{2\eta_t} \frac{1}{|\mathcal{S}|} \sum_{k \in \mathcal{S}} \Vert \mathbf{w}_k^t  - \mathbf{w}^t\Vert^2 < \frac{1}{|\mathcal{S}|} \sum_{k \in \mathcal{S}} \Vert \mathbf{w}_k^t  - \mathbf{w}^t\Vert^2$ and $[\gamma_t \eta_t L +  2L\eta_t^2 ( |\mathcal{I}| \psi + |\mathcal{S}|) ]\Lambda < 2L\eta_t^2 ( |\mathcal{I}|\psi+|\mathcal{S}|+\varepsilon ) \Lambda$. 

By replacing term $\mathfrak{C}_1$ in (\ref{B10}) with (\ref{B13}), taking the expectation on both sides of (\ref{B10}) and leveraging Lemma \ref{lem1} to represent $\mathbb{E}[\mathfrak{C}_1]$, we have 
\begin{align*}
\label{B15}
& \mathbb{E} \Vert \mathbf{w}^{t+1} - \mathbf{w}^* \Vert^2 \\
\leq  & (1-\eta_t \mu \varepsilon ) \mathbb{E} \Vert \mathbf{w}^{t} - \mathbf{w}^{*} \Vert^2 + \mathbb{E} \sum_{k\in \mathcal{S}} \Vert \mathbf{w}^{t} - \mathbf{w}_k^{t} \Vert^2  \\ & +  \mathbb{E} \frac{1}{|\mathcal{S}|} \sum_{k \in \mathcal{S}} \Vert \mathbf{w}_k^t - \mathbf{w}^t \Vert^2 +  2L\eta_t^2 ( |\mathcal{I}|\psi+|\mathcal{S}|+\varepsilon ) \Lambda \\ & + \mathbb{E} [\eta_t^2 \mathfrak{C}_1] + \mathbb{E}[\mathfrak{C}_2].
\tag{C13}
\end{align*}

$\bullet$ \noindent Bounding term $\mathbb{E} \frac{1}{|\mathcal{S}|} \sum_{k \in \mathcal{S}} \Vert \mathbf{w}_k^t - \mathbf{w}^t \Vert^2$\\

Assume that between any two consecutive rounds, there is an aggregated model $\mathbf{w}^{t-1,r}, 1 \leq r \leq \tau$, which is not achieved in reality since aggregation happens only after every $\tau$ local steps. It is straightforward that $\mathbf{w}^{t-1,\tau} = \mathbf{w}^{t}$. The learning rate $\eta$ is fixed between two consecutive rounds. With that, $\mathbb{E} \sum_{k\in \mathcal{S}} \frac{1}{|\mathcal{S}|} \Vert \mathbf{w}^{t} - \mathbf{w}_k^{t} \Vert^2$ is bounded as follows
\begin{align*}
\label{D1}
& \mathbb{E} \sum_{k\in \mathcal{S}} \frac{1}{|\mathcal{S}|} \Vert \mathbf{w}^{t} - \mathbf{w}_k^{t} \Vert^2 \\ =& \, \mathbb{E} \sum_{k\in \mathcal{S}} \frac{1}{|\mathcal{S}|} \Vert (\mathbf{w}_k^{t} - \mathbf{w}^{t-1,r}) - ( \mathbf{w}^{t} - \mathbf{w}^{t-1,r} ) \Vert^2 \\ 
\stackrel{7} \leq &  \, \mathbb{E} \sum_{k\in \mathcal{S}} \frac{1}{|\mathcal{S}|} \Vert \mathbf{w}_k^{t} - \mathbf{w}^{t-1,r} \Vert^2 \\ 
 \stackrel{8}\leq &  \, \sum_{k\in \mathcal{S}}\frac{1}{|\mathcal{S}|} \mathbb{E} \sum_{j=r}^{\tau} (\tau-r) \eta_{t-1}^2 \Vert \nabla F_k(\mathbf{w}_{k}^{t-1,j}, \xi_k)  \Vert^2 
\\  \stackrel{9}\leq &  \, \sum_{k\in \mathcal{S}} \frac{1}{|\mathcal{S}|} \sum_{j=r}^{\tau} (\tau-r) \eta_{t-1}^2 G^2 
\\ \leq & \, \sum_{k\in \mathcal{S}} \frac{1}{|\mathcal{S}|} (\tau-1)^2 \eta_{t-1}^2 G^2 
\\  \stackrel{10} \leq & \, 4 \eta_t^2 (\tau-1)^2G^2 
\tag{C14} 
\end{align*} 
where the inequality $7$ is from $ \mathbb{E} \Vert X - \mathbb{E} X \Vert^2 \leq \mathbb{E} \Vert X \Vert^2$ \cite{PNS} and the inequality $8$ is achieved by Jensen inequality $\Vert \mathbf{w}_k^{t} - \mathbf{w}^{t-1,r} \Vert^2 = \Vert  \sum_{j=r}^{\tau} \eta_{t-1} \nabla F_k(\mathbf{w}_{k}^{t-1,j}, \xi_k)  \Vert^2 \leq (\tau -r) \sum_{j=r}^{\tau} \eta_{t-1}^2 \Vert \nabla F_k(\mathbf{w}_{k}^{t-1,j}, \xi_k)  \Vert^2$. Inequality $9$ is from Assumption 3, and the inequality $10$ holds since $\eta_{t-1} \leq 2  \eta_{t}$.

Analogously, we can bound $\mathbb{E} \sum_{k\in \mathcal{S}} \Vert \mathbf{w}^{t} - \mathbf{w}_k^{t} \Vert^2 \leq 4 \eta_t^2 (\tau-1)^2G^2$ in the same way.  

By inserting (\ref{D1}) to (\ref{B15}), we have
\begin{align*}
& \mathbb{E} \Vert \mathbf{w}^{t+1} - \mathbf{w}^* \Vert^2 \\
\leq  & (1-\eta_t \mu  \varepsilon ) \mathbb{E} \Vert \mathbf{w}^{t} - \mathbf{w}^{*} \Vert^2 + 8\eta_t^2 (\tau-1)^2G^2 \\& +2L\eta_t^2 ( |\mathcal{I}|\psi+|\mathcal{S}|+\varepsilon ) \Lambda  + 2 \eta_t^2 \delta^2 \psi,
\tag{C15}
\end{align*}
where the inequality holds because $\mathbb{E}[\mathfrak{C}_2] = \mathbb{E}[2\eta_t \langle \mathbf{w}^{t}- \eta_t \nabla \bar F(\mathbf{w}^{t}) - \mathbf{w}^{*}, \nabla \bar F(\mathbf{w}^{t}) - \nabla F(\mathbf{w}^{t}) \rangle] = 0$ due to $\mathbb{E}[ \nabla F(\mathbf{w}^{t})] = \nabla \bar F(\mathbf{w}^{t}) $ and $\mathbb{E}[\mathfrak{C}_1] = 2 \eta_t^2 \delta^2 \psi$ by Lemma \ref{lem1}. \\

\subsection{Proof of Theorem 1}
\label{Proof-Theorem}
From Lemma \ref{lem2}, it follows that $\Gamma_{t+1} \leq (1-\eta_t \mu \varepsilon ) \Gamma_{t} + \eta_t^2 \Delta$
where $\Gamma_{t+1} = \mathbb{E} \left[ \Vert \mathbf{w}^{t+1} - \mathbf{w}^{*} \Vert^2 \right]$, $\Gamma_{t} = \mathbb{E} \left[ \Vert \mathbf{w}^{t} - \mathbf{w}^{*} \Vert^2 \right]$ and $\Delta = 8 (\tau-1)^2G^2 + 2L ( |\mathcal{I}|\psi+|\mathcal{S}|+\varepsilon ) \Lambda  + 2 \delta^2 \psi$.

For a diminishing step size $\eta_t = \frac{\beta}{t+\lambda}$ and for some $\lambda>0, \beta >\frac{1}{\mu}$ such that $\eta_t \leq \frac{\varepsilon}{2L (|\mathcal{I}| \psi + |\mathcal{S}|) }$ and  $\eta_t \leq 2\eta_{t+1}$, we aim to prove $\Gamma_t \leq \frac{\upsilon}{t+\lambda}$ where $\upsilon = \max\{(\lambda +1 )\Gamma_1, \frac{ \beta^2 \Delta}{ \beta \mu \varepsilon -1 }\}$. 

Firstly, the definition of $\upsilon$ ensures that $\Gamma_t$ holds for $t=1$. Assume that $\Gamma_t \leq \frac{\upsilon}{t+\lambda}$ holds for some $t$, we have
\begin{align*}
& \Gamma_{t+1} \leq (1-\eta_t \mu \varepsilon ) \Gamma_{t} + \eta_t^2 \Delta  \\
& \leq (1 - \frac{\beta \mu \varepsilon}{t + \lambda}) \frac{\upsilon}{t+\lambda} + \frac{\beta^2 \Delta }{(t+\lambda)^2} \\
& =   \frac{ t+\lambda -1 } {(t+\lambda)^2} \upsilon + [ \frac{ \beta^2 \Delta } {(t+\lambda)^2} - \frac{ \beta \mu \varepsilon -1 } {(t+\lambda)^2} \upsilon ]\\
&  \leq \frac{\upsilon}{t+\lambda+1} .
\end{align*}

By the definition of $\upsilon$,
\begin{align*}
    \upsilon & =  \max\{ \frac{ \beta^2 \Delta}{ \beta \mu \varepsilon -1},  (\lambda +1 )\Gamma_1  \} \leq  (\lambda +1 )\Gamma_1 + \frac{ \beta^2 \Delta}{ \beta \mu \varepsilon  -1}
\end{align*}

Then, by choosing $\beta = \frac{2}{\mu \varepsilon}$ ($\eta_t = \frac{2}{\mu \varepsilon (t+\lambda)}$ in the meantime) and using $L$-smoothness property of $F$, Theorem \ref{thm1} is proven as 
\begin{align*}
& \mathbb{E}\left[ F(\mathbf{w}^{T})  - F(\mathbf{w}^{*}) \right] \leq \frac{L}{2} \Gamma_T \\ & \leq \frac{1}{T+\lambda} \left( \frac{(\lambda +1 )\Gamma_1}{2} + \frac{ 2 \Delta}{ \mu^2 \varepsilon^2}   \right)
\end{align*}

\bibliographystyle{IEEEtran}
\bibliography{reference}

\begin{thebibliography}{10}
\providecommand{\url}[1]{#1}
\csname url@samestyle\endcsname
\providecommand{\newblock}{\relax}
\providecommand{\bibinfo}[2]{#2}
\providecommand{\BIBentrySTDinterwordspacing}{\spaceskip=0pt\relax}
\providecommand{\BIBentryALTinterwordstretchfactor}{4}
\providecommand{\BIBentryALTinterwordspacing}{\spaceskip=\fontdimen2\font plus
\BIBentryALTinterwordstretchfactor\fontdimen3\font minus
  \fontdimen4\font\relax}
\providecommand{\BIBforeignlanguage}[2]{{%
\expandafter\ifx\csname l@#1\endcsname\relax
\typeout{** WARNING: IEEEtran.bst: No hyphenation pattern has been}%
\typeout{** loaded for the language `#1'. Using the pattern for}%
\typeout{** the default language instead.}%
\else
\language=\csname l@#1\endcsname
\fi
#2}}
\providecommand{\BIBdecl}{\relax}
\BIBdecl

\bibitem{10233400}
H.~Wu, P.~Wang, and A.~C. Narayan, ``Model-heterogeneous federated learning
  with partial model training,'' in \emph{Proc. IEEE/CIC International
  Conference on Communications in China (ICCC)}, 2023.

\bibitem{FedAvg}
B.~McMahan, E.~Moore, D.~Ramage, S.~Hampson, and B.~A. y~Arcas,
  ``Communication-efficient learning of deep networks from decentralized
  data,'' in \emph{Proc. the Artificial Intelligence and Statistics Conference
  (AISTATS)}, 2017.

\bibitem{park2019wireless}
J.~Park, S.~Samarakoon, M.~Bennis, and M.~Debbah, ``Wireless network
  intelligence at the edge,'' \emph{Proceedings of the IEEE}, vol. 107, no.~11,
  pp. 2204--2239, 2019.

\bibitem{quantizatino}
S.~Zheng, C.~Shen, and X.~Chen, ``Design and analysis of uplink and downlink
  communications for federated learning,'' \emph{IEEE Journal on Selected Areas
  in Communications}, vol.~39, no.~7, pp. 2150--2167, 2021.

\bibitem{compression}
S.~P. Karimireddy, Q.~Rebjock, S.~Stich, and M.~Jaggi, ``Error feedback fixes
  signsgd and other gradient compression schemes,'' in \emph{Proc.
  International Conference on Machine Learning (ICML)}, 2019.

\bibitem{sparsification}
P.~Han, S.~Wang, and K.~K. Leung, ``Adaptive gradient sparsification for
  efficient federated learning: An online learning approach,'' in \emph{Proc.
  IEEE International Conference on Distributed Computing Systems (ICDCS)},
  2020.

\bibitem{overtheair}
K.~Yang, T.~Jiang, Y.~Shi, and Z.~Ding, ``Federated learning via over-the-air
  computation,'' \emph{IEEE Transactions on Wireless Communications}, vol.~19,
  no.~3, pp. 2022--2035, 2020.

\bibitem{scaffold}
S.~P. Karimireddy, S.~Kale, M.~Mohri, S.~Reddi, S.~Stich, and A.~T. Suresh,
  ``Scaffold: Stochastic controlled averaging for federated learning,'' in
  \emph{Proc. International Conference on Machine Learning (ICML)}, 2020.

\bibitem{BiasedSelection}
Y.~Jee~Cho, J.~Wang, and G.~Joshi, ``Towards understanding biased client
  selection in federated learning,'' in \emph{Proc. the International
  Conference on Artificial Intelligence and Statistics (AISTATS)}, 2022.

\bibitem{PNS}
H.~Wu and P.~Wang, ``Node selection toward faster convergence for federated
  learning on non-iid data,'' \emph{IEEE Transactions on Network Science and
  Engineering}, vol.~9, no.~5, pp. 3099--3111, 2022.

\bibitem{9796935}
B.~Luo, W.~Xiao, S.~Wang, J.~Huang, and L.~Tassiulas, ``Tackling system and
  statistical heterogeneity for federated learning with adaptive client
  sampling,'' in \emph{Proc. IEEE Conference on Computer Communications
  (INFOCOM)}, 2022.

\bibitem{Adp}
H.~Wu and P.~Wang, ``Fast-convergent federated learning with adaptive
  weighting,'' \emph{IEEE Transactions on Cognitive Communications and
  Networking}, vol.~7, no.~4, pp. 1078--1088, 2021.

\bibitem{FirstorderP}
M.~Zhang, K.~Sapra, S.~Fidler, S.~Yeung, and J.~M. Alvarez, ``Personalized
  federated learning with first order model optimization,'' in \emph{Proc.
  International Conference on Learning Representations (ICML)}, 2021.

\bibitem{SPcoding}
H.~Baek, W.~J. Yun, Y.~Kwak, S.~Jung, M.~Ji, M.~Bennis, J.~Park, and J.~Kim,
  ``Joint superposition coding and training for federated learning over
  multi-width neural networks,'' in \emph{Proc. IEEE Conference on Computer
  Communications (INFOCOM)}, 2022.

\bibitem{briggs2020federated}
C.~Briggs, Z.~Fan, and P.~Andras, ``Federated learning with hierarchical
  clustering of local updates to improve training on non-iid data,'' in
  \emph{2020 International Joint Conference on Neural Networks (IJCNN)}.\hskip
  1em plus 0.5em minus 0.4em\relax IEEE, 2020, pp. 1--9.

\bibitem{IFCA}
A.~Ghosh, J.~Chung, D.~Yin, and K.~Ramchandran, ``An efficient framework for
  clustered federated learning,'' in \emph{Proc. Advances in Neural Information
  Processing Systems (NeurIPS)}, 2020.

\bibitem{nishio2019client}
T.~Nishio and R.~Yonetani, ``Client selection for federated learning with
  heterogeneous resources in mobile edge,'' in \emph{Proc. International
  Conference on Communications (ICC)}, 2019.

\bibitem{9488679}
B.~Luo, X.~Li, S.~Wang, J.~Huang, and L.~Tassiulas, ``Cost-effective federated
  learning design,'' in \emph{Proc. IEEE Conference on Computer Communications
  (INFOCOM)}, 2021.

\bibitem{9261995}
C.~T. Dinh, N.~H. Tran, M.~N.~H. Nguyen, C.~S. Hong, W.~Bao, A.~Y. Zomaya, and
  V.~Gramoli, ``Federated learning over wireless networks: Convergence analysis
  and resource allocation,'' \emph{IEEE/ACM Transactions on Networking},
  vol.~29, no.~1, pp. 398--409, 2021.

\bibitem{amiri2021convergence}
M.~M. Amiri, D.~G{\"u}nd{\"u}z, S.~R. Kulkarni, and H.~V. Poor, ``Convergence
  of update aware device scheduling for federated learning at the wireless
  edge,'' \emph{IEEE Transactions on Wireless Communications}, vol.~20, no.~6,
  pp. 3643--3658, 2021.

\bibitem{chen2021communication}
M.~Chen, N.~Shlezinger, H.~V. Poor, Y.~C. Eldar, and S.~Cui,
  ``Communication-efficient federated learning,'' \emph{Proceedings of the
  National Academy of Sciences}, vol. 118, no.~17, p. e2024789118, 2021.

\bibitem{bommasani2021opportunities}
R.~Bommasani, D.~A. Hudson, E.~Adeli, R.~Altman, S.~Arora, S.~von Arx, M.~S.
  Bernstein, J.~Bohg, A.~Bosselut, E.~Brunskill \emph{et~al.}, ``On the
  opportunities and risks of foundation models,'' \emph{arXiv preprint
  arXiv:2108.07258}, 2021.

\bibitem{resnet}
K.~He, X.~Zhang, S.~Ren, and J.~Sun, ``Deep residual learning for image
  recognition,'' in \emph{Proc. IEEE Conference on Computer Vision and Pattern
  Recognition (CVPR)}, 2016.

\bibitem{transformer}
A.~Vaswani, N.~Shazeer, N.~Parmar, J.~Uszkoreit, L.~Jones, A.~N. Gomez,
  {\L}.~Kaiser, and I.~Polosukhin, ``Attention is all you need,'' in
  \emph{Proc. Advances in neural information processing systems (NeurIPS)},
  2017.

\bibitem{hinton2015distilling}
G.~Hinton, O.~Vinyals, J.~Dean \emph{et~al.}, ``Distilling the knowledge in a
  neural network,'' \emph{arXiv preprint arXiv:1503.02531}, vol.~2, no.~7,
  2015.

\bibitem{lin2020ensemble}
T.~Lin, L.~Kong, S.~U. Stich, and M.~Jaggi, ``Ensemble distillation for robust
  model fusion in federated learning,'' \emph{Advances in Neural Information
  Processing Systems}, vol.~33, pp. 2351--2363, 2020.

\bibitem{cho2021personalized}
Y.~J. Cho, J.~Wang, T.~Chirvolu, and G.~Joshi, ``Communication-efficient and
  model-heterogeneous personalized federated learning via clustered knowledge
  transfer,'' \emph{IEEE Journal of Selected Topics in Signal Processing}, pp.
  1--14, 2023.

\bibitem{he2020group}
C.~He, M.~Annavaram, and S.~Avestimehr, ``Group knowledge transfer: Federated
  learning of large cnns at the edge,'' in \emph{Proc. Advances in Neural
  Information Processing Systems (NeurIPS)}, vol.~33, 2020, pp.
  14\,068--14\,080.

\bibitem{FedGen}
Z.~Zhu, J.~Hong, and J.~Zhou, ``Data-free knowledge distillation for
  heterogeneous federated learning,'' in \emph{Proc. International Conference
  on Machine Learning (ICML)}, 2021.

\bibitem{FedMP}
Z.~Jiang, Y.~Xu, H.~Xu, Z.~Wang, C.~Qiao, and Y.~Zhao, ``Fedmp: Federated
  learning through adaptive model pruning in heterogeneous edge computing,'' in
  \emph{2022 IEEE 38th International Conference on Data Engineering (ICDE)},
  2022, pp. 767--779.

\bibitem{JiangIBM}
Y.~Jiang, S.~Wang, V.~Valls, B.~J. Ko, W.-H. Lee, K.~K. Leung, and
  L.~Tassiulas, ``Model pruning enables efficient federated learning on edge
  devices,'' \emph{IEEE Transactions on Neural Networks and Learning Systems},
  pp. 1--13, 2022.

\bibitem{9835327}
Z.~Jiang, Y.~Xu, H.~Xu, Z.~Wang, C.~Qiao, and Y.~Zhao, ``Fedmp: Federated
  learning through adaptive model pruning in heterogeneous edge computing,'' in
  \emph{Proc. International Conference on Data Engineering (ICDE)}, 2022.

\bibitem{dropout}
D.~Wen, K.-J. Jeon, and K.~Huang, ``Federated dropout—a simple approach for
  enabling federated learning on resource constrained devices,'' \emph{IEEE
  Wireless Communications Letters}, vol.~11, no.~5, pp. 923--927, 2022.

\bibitem{diao2021heterofl}
E.~Diao, J.~Ding, and V.~Tarokh, ``Heterofl: Computation and communication
  efficient federated learning for heterogeneous clients,'' in \emph{Proc.
  International Conference on Learning Representations (ICLR)}, 2021.

\bibitem{fjord}
S.~Horvath, S.~Laskaridis, M.~Almeida, I.~Leontiadis, S.~Venieris, and N.~Lane,
  ``Fjord: Fair and accurate federated learning under heterogeneous targets
  with ordered dropout,'' in \emph{Proc. Advances in Neural Information
  Processing Systems (NeurIPS)}, 2021.

\bibitem{FedRolex}
S.~Alam, L.~Liu, M.~Yan, and M.~Zhang, ``Fedrolex: Model-heterogeneous
  federated learning with rolling sub-model extraction,'' in \emph{Proc.
  Advances in Neural Information Processing Systems (NeurIPS)}, 2022.

\bibitem{yu2018slimmable}
J.~Yu, L.~Yang, N.~Xu, J.~Yang, and T.~Huang, ``Slimmable neural networks,''
  \emph{arXiv preprint arXiv:1812.08928}, 2018.

\bibitem{yu2019universally}
J.~Yu and T.~S. Huang, ``Universally slimmable networks and improved training
  techniques,'' in \emph{Proc. IEEE/CVF International Conference on Computer
  Vision (ICCV)}, 2019.

\bibitem{yuan2019distributed}
B.~Yuan, C.~R. Wolfe, C.~Dun, Y.~Tang, A.~Kyrillidis, and C.~M. Jermaine,
  ``Distributed learning of deep neural networks using independent subnet
  training,'' \emph{arXiv preprint arXiv:1910.02120}, 2019.

\bibitem{ClassifierCalibration}
M.~Luo, F.~Chen, D.~Hu, Y.~Zhang, J.~Liang, and J.~Feng, ``No fear of
  heterogeneity: Classifier calibration for federated learning with non-iid
  data,'' in \emph{Proc. Advances in Neural Information Processing Systems
  (NeurIPS)}, 2021.

\bibitem{ANIL}
A.~Raghu, M.~Raghu, S.~Bengio, and O.~Vinyals, ``Rapid learning or feature
  reuse? towards understanding the effectiveness of maml,'' in \emph{Proc.
  International Conference on Learning Representations (ICLR)}, 2020.

\bibitem{sun2017meprop}
X.~Sun, X.~Ren, S.~Ma, and H.~Wang, ``meprop: Sparsified back propagation for
  accelerated deep learning with reduced overfitting,'' in \emph{Proc.
  International Conference on Machine Learning (ICML)}, 2017.

\bibitem{mask_convergent}
A.~Mohtashami, M.~Jaggi, and S.~Stich, ``Masked training of neural networks
  with partial gradients,'' in \emph{Proc. International Conference on
  Artificial Intelligence and Statistics (AISTATS)}, 2022.

\bibitem{Li2020On}
X.~Li, K.~Huang, W.~Yang, S.~Wang, and Z.~Zhang, ``On the convergence of fedavg
  on non-iid data,'' in \emph{Proc. International Conference on Learning
  Representations (ICLR)}, 2020.

\bibitem{haddadpour2019convergence}
F.~Haddadpour and M.~Mahdavi, ``On the convergence of local descent methods in
  federated learning,'' \emph{arXiv preprint arXiv:1910.14425}, 2019.

\bibitem{9139873}
Y.~Zhan, P.~Li, and S.~Guo, ``Experience-driven computational resource
  allocation of federated learning by deep reinforcement learning,'' in
  \emph{Proc. IEEE International Parallel and Distributed Processing Symposium
  (IPDPS)}, 2020.

\bibitem{8664630}
S.~Wang, T.~Tuor, T.~Salonidis, K.~K. Leung, C.~Makaya, T.~He, and K.~Chan,
  ``Adaptive federated learning in resource constrained edge computing
  systems,'' \emph{IEEE Journal on Selected Areas in Communications}, vol.~37,
  no.~6, pp. 1205--1221, 2019.

\bibitem{stich2018local}
S.~U. Stich, ``Local sgd converges fast and communicates little,'' \emph{arXiv
  preprint arXiv:1805.09767}, 2018.

\bibitem{fedprox}
T.~Li, A.~K. Sahu, M.~Zaheer, M.~Sanjabi, A.~Talwalkar, and V.~Smith,
  ``Federated optimization in heterogeneous networks,'' \emph{Proc. Machine
  Learning and Systems (MLSys)}, 2020.

\bibitem{he2015convolutional}
K.~He and J.~Sun, ``Convolutional neural networks at constrained time cost,''
  in \emph{Proc. IEEE conference on computer vision and pattern recognition
  (CVPR)}, 2015, pp. 5353--5360.

\end{thebibliography}



\end{document}